%% file: ms.tex
\documentclass{article}

\usepackage{arxiv}

\usepackage[utf8]{inputenc} % allow utf-8 input
\usepackage[T1]{fontenc}    % use 8-bit T1 fonts

\usepackage{xr-hyper-mod}  %V: to enable crosslinks - using modified version that uses the extension formerlyaux instead of just plain aux as arxiv removes the aux files
\usepackage{hyperref}       % hyperlinks
\usepackage{url}            % simple URL typesetting
\usepackage{booktabs}       % professional-quality tables
\usepackage{amsfonts}       % blackboard math symbols
\usepackage{nicefrac}       % compact symbols for 1/2, etc.
\usepackage{microtype}      % microtypography
\usepackage{lipsum}         % Can be removed after putting your text content
\usepackage[pdftex]{graphicx}
\usepackage{textcase}

\usepackage{amssymb}
\usepackage{amsfonts}
\usepackage{mathtools}

\PassOptionsToPackage{american}{babel} % change this to your language(s), main language last
    \usepackage{babel}

\usepackage{csquotes}
\PassOptionsToPackage{%
  backend=biber,bibencoding=utf8, %instead of bibtex
  language=auto,%
  style=numeric-comp,%
  giveninits=true,  % V: show only initials
  sorting=none,
  maxbibnames=10, % default: 3, et al.
  backref=true,%
  natbib=true % natbib compatibility mode (\citep and \citet still work)
}{biblatex}
    \usepackage{biblatex}

\PassOptionsToPackage{printonlyused,smaller}{acronym}
	\usepackage{acronym} % nice macros for handling all acronyms in the thesis

\usepackage[version=4]{mhchem}

\usepackage{array}
\newcolumntype{T}{>{\tiny}l} % define a new column type for \tiny

\usepackage{amssymb}% http://ctan.org/pkg/amssymb
\usepackage{pifont}% http://ctan.org/pkg/pifont
\newcommand{\cmark}{\ding{51}}%
\newcommand{\xmark}{\ding{55}}%

\usepackage{physics}

\usepackage{bm}

\usepackage{listings}

\usepackage{xcolor}

\definecolor{codegreen}{rgb}{0,0.6,0}
\definecolor{codegray}{rgb}{0.5,0.5,0.5}
\definecolor{codepurple}{rgb}{0.58,0,0.82}
\definecolor{backcolour}{rgb}{0.95,0.95,0.92}

\usepackage{color}
\usepackage{nth}

\usepackage{xurl}

\usepackage[T1]{fontenc}

\usepackage{caption}
\usepackage{subcaption}
\usepackage[space]{grffile}

\usepackage{float}
\usepackage{longtable}            
\usepackage{rotating}
\usepackage{pdflscape}
\usepackage{flafter}
\usepackage{afterpage}
\usepackage{placeins}
\usepackage{geometry}

\usepackage[english]{isodate}

\usepackage[acronyms,toc=false]{glossaries-extra}
\makeglossaries
\glsdisablehyper

\newcommand{\seeref}[2]{{#1}, \emph{see \cref{#2}};}

\newcommand{\myTitle}{Exploring the Relationship: Transformative Adaptive Activation Functions in Comparison to Other Activation Functions}
\newcommand{\myKeywords}{transformative adaptive activation functions, activation functions, neural network}

\usepackage{cleveref}       % smart cross-referencing
\usepackage{kvsetkeys}

\makeatletter
\let\org@@cref\@cref
\renewcommand*{\@cref}[2]{%
  \edef\process@me{%
    \noexpand\org@@cref{#1}{\zap@space#2 \@empty}%
  }\process@me
}
\makeatother

\DeclareUnicodeCharacter{3B1}{\ensuremath{\alpha}}

\title{\myTitle}

\newif\ifuniqueAffiliation
\uniqueAffiliationtrue

\ifuniqueAffiliation % Standard variant of author block
\author{ \href{https://orcid.org/0000-0002-5130-4384}{\includegraphics[scale=0.06]{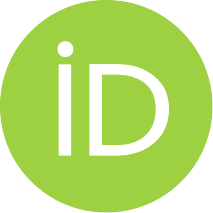}\hspace{1mm}Vladimír Kunc}
\\
\href{https://cs.fel.cvut.cz/}{\color{black}{Department of Computer Science}}\\
\href{https://fel.cvut.cz/}{\color{black}{Faculty of Electrical Engineering}}\\
		\href{https://www.cvut.cz/}{\color{black}{Czech Technical University in Prague}}\\
	\texttt{kuncvlad@fel.cvut.cz} \\
}
\else
\usepackage{authblk}

\newbox{\orcid}\sbox{\orcid}{\includegraphics[scale=0.06]{orcid.pdf}} 
\author[1]{%
	\href{https://orcid.org/0000-0002-5130-4384}{\usebox{\orcid}\hspace{1mm}Vladimír Kunc\thanks{\texttt{kuncvlad@fel.cvut.cz}}}%
}
\affil[1]{\href{https://cs.fel.cvut.cz/}{\color{black}{Department of Computer Science}}, \href{https://fel.cvut.cz/}{\color{black}{Faculty of Electrical Engineering}}, \href{https://www.cvut.cz/}{\color{black}{Czech Technical University in Prague}}}
\fi

\renewcommand{\headeright}{Technical Report}
\renewcommand{\undertitle}{Technical Report}
\renewcommand{\shorttitle}{TAAFs in Comparison to Other Activation Functions}

\hypersetup{
pdftitle={\myTitle},
pdfauthor={Vladimír Kunc, Jiří Kléma},
pdfkeywords={\myKeywords},
}

\input{tech_acronyms.tex}
\begin{document}
\maketitle

\begin{abstract}
Neural networks are the state-of-the-art approach for many tasks and the activation function is one of the main building blocks that allow such performance. Recently, a novel transformative adaptive activation function (TAAF) allowing for any vertical and horizontal translation and scaling was proposed. This work sets the TAAF into the context of other activation functions. It shows that the TAAFs generalize over 50 existing activation functions and utilize similar concepts as over 70 other activation functions, underscoring the versatility of TAAFs. This comprehensive exploration positions TAAFs as a promising and adaptable addition to neural networks.

\end{abstract}
% keywords can be removed

\section{Introduction}
\Glspl{NN} represent the state-of-the-art in many contermporary challenges and are able to solve various tasks ranging from image classification (e.g., \cite{He2015,Szegedy2015,Szegedy2017,Maurcio2023,Chen2021Review}) over prediction of time-series (e.g., \cite{Torres2021,Tealab2018}) to natural language processing (e.g., \cite{Devlin2019,Touvron2023,Zhou2020Progress},) and art generation (e.g., \cite{Santos2021,Maerten2023}).\footnote{More reviews and examples of \glspl{NN} solving various tasks are available in books and reviews such as \cite{Goodfellow2016,Aggarwal2023, Schmidhuber2015, Pouyanfar2018, Khan2020, LeCun2015, Guo2016, Shrestha2019,Rawat2017, Min2016, Angermueller2016, Zamani2013, Alom2019, Zhou2020Graph, Bacciu2020, Guo2021, Wang2021Deep, Tian2020, Litjens2017, Fourcade2019, Zhao2019Object, Sze2017, Kamilaris2018, Zhang2018ASurveyOnDeep, Meyer2018, Yamashita2018, Bronstein2017, Voulodimos2018, Arulkumaran2017, Abiodun2018,  Karniadakis2021, Lundervold2019, Chen2018TheRise, Lee2017, Akkus2017, Chen2019Deep, Anwar2018, Amin2021, Razzak2017, Suganyadevi2021, Helaly2023, SudheerKumar2019, Hafiz2019, Liu2022Deep, Qiu2020, Moen2019, Jing2021, Yang2019, Eraslan2019, Tavanaei2019, Wang2019AReview, Goh2017, Ball2017, vanKlompenburg2020, Sahiner2018}.} One of the main building blocks of a \gls{NN} is the \gls[prereset]{AF} that introduces non-linearity to the computation \cite{Dubey2022}.

\Gls[prereset]{AAF} are a class of \glspl{AF} that are able to adapt to the data --- usually by introducing a parameter controlling properties of the \gls{AF} that is trained together with the network's weights \cite{Apicella2021}. These \glspl{AAF} are getting popular as they often lead to better network performance \cite{Apicella2021}. There are many fixed and adaptive \glspl{AF} --- over 400 various \glspl{AF} were proposed in the literature \cite{Kunc2024Decades} --- but this work focuses on the \gls[prereset]{TAAF}. The \gls{TAAF} is a family of \glspl{AF} proposed in \cite{Kunc2021, Kunc2020} that allows for scaling and translation of any inner activation functions using four additional parameters. Despite the simplicity, the \gls{TAAF} can be used to model many adaptive and nonadaptive \glspl{AF} proposed in the literature. The goal of this work is to show the special cases of the \glspl{TAAF} that were proposed in the literature and to discuss \glspl{AF} that utilize similar concepts --- references to sections in \cite{Kunc2024Decades} will be provided to save space and avoid a redundant list of definitions \glspl{AF}.

\section{Transformative adaptive activation function}
\label{sec:taaf_local_definition}
As stated above, the \gls{TAAF} \cite{Kunc2021,Kunc2020} introduce four adaptive parameters; this allows for horizontal and vertical scaling and translation of any inner \gls{AF}. The \gls{TAAF} is defined as
\begin{equation}
    g(f,z_i) = \alpha_i \cdot f(\beta_i\cdot z_i + \gamma_i) +\delta_i,
\end{equation}
where $z_i$ input to the \gls{AF}, $\alpha_i$, $\beta_i$, $\gamma_i$, and $\delta_i$, are trainable parameters for each neuron $i$ \cite{Kunc2021,Kunc2024Decades}. The output of a neuron with \gls{TAAF} with inputs $x_i$ is:
\begin{equation}
   \alpha_i \cdot f\left(\beta_i\cdot \sum_{i=1}^nw_ix_i + \gamma_i\right) +\delta_i,
\end{equation}
where $x_i$ are individual inputs, $w_i$ are its weights, and $n$ is the number of incoming connections \cite{Kunc2021,Kunc2024Decades}.

\input{section_motivation.tex}

\section{Conclusion}
\label{sec:conclusion}

The \glspl{TAAF} generalize over 50 \glspl{AF} proposed in the literature that can be considered special cases of the \glspl{TAAF} --- with some of the four adaptive parameters fixed and not trainable. Furthermore, over 70 \glspl{AF} utilize similar concepts and ideas as are behind the adaptive parameters of the \glspl{TAAF}. This confirms the motivations behind the four adaptive parameters of the \glspl{TAAF} and since each of the parameters was used in an \gls{AF} proposed in the literature, it also adds further evidence that the parameters were correctly chosen. This work sets the \glspl{TAAF} to the context of contemporary \glsxtrlongpl{AF}.

\defbibheading{bibintoc}[References]{%
  \phantomsection
  \pdfbookmark[0]{#1}{#1}
  \section*{#1}%
}
\AtNextBibliography{\scriptsize}
\printbibliography[heading=bibintoc]

\end{document}

%% file: tech_acronyms.tex
\setabbreviationstyle[acronym]{long-short}
\newcommand{\newglossaryentrywithacronym}[4]{
    \newglossaryentry{#2_gls}{
        name={#2},
        long={#3},
        description={#4}
    }

    \newglossaryentry{#1}{
        type=\acronymtype,
        name={#1},
        short={{#1}\glsadd{#2_gls}},
        text={{#1}\glsadd{#2_gls}},
        long={{#3}\glsadd{#2_gls}},
        shortplural={{#1s}\glsadd{#2_gls}},
        longplural={{#3s}\glsadd{#2_gls}},
        description={#3},
        first={#3 (#1)\glsadd{#2_gls}},
        firstplural={#3s (#1s)\glsadd{#2_gls}},
        see=[Glossary:]{#2_gls}
    }

}
\newcommand{\newacronymwithref}[3]{
    \newglossaryentry{#1}{
        type=\acronymtype,
        name={#1},
        short={#1},
        long={#2},
        first={#2 (#1)},
        shortplural={#1s},
        longplural={#2s},
        firstplural={#2s (#1s)},
        description={\seeref{#2}{#3}},
    }
}

\newcommand{\newacronymwithrefdesc}[4]{
    \newglossaryentry{#1}{
        type=\acronymtype,
        name={#1},
        short={#1},
        long={#2},
        first={#2 (#1)},
        shortplural={#1s},
        longplural={#2s},
        firstplural={#2s (#1s)},
        description={\seeref{#3}{#4}},
    }
}

\newcommand{\newgacronymwithdesc}[3]{
    \newglossaryentry{#1}{
        type=\acronymtype,
        name={#1},
        short={#1},
        long={#2},
        first={#2 (#1)},
        shortplural={#1s},
        longplural={#2s},
        firstplural={#2s (#1s)},
        description={#3},
    }
}

\newcommand{\newacronymwithcustomshortdesc}[4]{
    \newglossaryentry{#1}{
        type=\acronymtype,
        name={#2},
        short={#2},
        long={#3},
        first={#3 (#2)},
        shortplural={#2s},
        longplural={#3s},
        firstplural={#3s (#2s)},
        description={#4},
    }
}

\newcommand{\newglossaryentrywithacronymandref}[5]{
    \newglossaryentry{#2_gls}{
        name={#2},
        long={#3},
        description={\seeref{#4}{#5}},
    }

    \newglossaryentry{#1}{
        type=\acronymtype,
        name={#1},
        short={{#1}\glsadd{#2_gls}},
        shortplural={{#1}s\glsadd{#2_gls}},
        text={{#1}\glsadd{#2_gls}},
        long={{#3}\glsadd{#2_gls}},
        longplural={{#3}s\glsadd{#2_gls}},
        description={\seeref{#3}{#5}},
        first={#3 (#1)\glsadd{#2_gls}},
        firstplural={#3s (#1s)\glsadd{#2_gls}},
        see=[Glossary:]{#2_gls}
    }
}

\begingroup
  \boolfalse{citerequest}

\newglossaryentry{WoS}{
    type=\acronymtype,
    name={WoS},
    short={WoS},
    long={Web of Science},
    description={\href{https://www.webofscience.com/wos}{Web of Science}},
    first={Web of Science (WoS)},
}

\newglossaryentrywithacronym{JIF}{Journal Impact Factor}{Journal Impact Factor}{a journal citation metric calculated by \href{https://clarivate.com/}{Clarivate} using items indexed in \glsxtrshort{WoS}; it is equal to the mean number of citations for articles published in the last two years\footnote{see \url{https://incites.help.clarivate.com/Content/Indicators-Handbook/ih-journal-citation-reports.htm}}}
\newglossaryentrywithacronym{JCI}{Journal Citation Indicator}{Journal Citation Indicator}{a journal citation metric calculated by \href{https://clarivate.com/}{Clarivate} using items indexed in \glsxtrshort{WoS}; it is equal to the average \glsxtrshort{CNCI} published in the last three years}
\newglossaryentrywithacronym{CNCI}{Category Normalized Citation Impact}{Category Normalized Citation Impact}{a document citation metric calculated by \href{https://clarivate.com/}{Clarivate}; it is equal to the number of a document's citations normalized by the expected number of citations for the same document type, publication year, and subject area \footnote{see \url{https://incites.help.clarivate.com/Content/Indicators-Handbook/ih-normalized-indicators.htm}}}

\newglossaryentry{CiteScore}{
    name={CiteScore},
    description={a journal citation metric calculated by \href{https://www.elsevier.com/}{Elsevier} using items indexed in \href{https://www.scopus.com/}{Scopus}; it is equal to the average number of citations per document published in the last four years \footnote{see \url{https://service.elsevier.com/app/answers/detail/a_id/14880/supporthub/scopus/}}},
}

\newglossaryentrywithacronym{A}{adenine}{adenine}{a purine nucleobase. One of the four bases in the \glsxtrshort{DNA}.}
\newglossaryentrywithacronym{G}{guanine}{guanine}{a purine nucleobase. One of the four bases in the \glsxtrshort{DNA}.}
\newglossaryentrywithacronym{C}{cytosine}{cytosine}{a pyrimidine nucleobase. One of the four bases in the \glsxtrshort{DNA}.}
\newglossaryentrywithacronym{T}{thymine}{thymine}{a pyrimidine nucleobase. One of the four bases in the \glsxtrshort{DNA}.}
\newglossaryentrywithacronym{DNA}{DNA}{deoxyribonucleic acid}{an extended molecular structure for storing hereditary information}
\newglossaryentrywithacronym{RNA}{RNA}{ribonucleic acid}{an extended molecular structure; stores hereditary information but also can catalyze biological reactions and control \glsxtrlong{GE} among other things}
\newglossaryentrywithacronym{GE}{gene expression}{gene expression}{process of synthesizing gene product (e.g., a protein) using information encoded in a gene}
\newglossaryentrywithacronym{cDNA}{cDNA}{copy DNA}{copy \glsxtrshort{DNA}, also called complementary \glsxtrshort{DNA}; synthetic DNA transcribed from \glsxtrshort{mRNA} \cite{Jaksik2015} using reverse transcriptase \cite[19]{BolonCanedo2019}}
\newglossaryentrywithacronym{cRNA}{cRNA}{copy RNA}{copy \glsxtrshort{RNA}, transcribed from \glsxtrshort{cDNA} during amplification phase}
\newglossaryentrywithacronym{mRNA}{messenger RNA}{messenger RNA}{copied from \glsxtrshort{DNA} during transcription; used for protein synthesis during translation}
\newglossaryentrywithacronym{miRNA}{microRNA}{microRNA}{a small, non-coding \glsxtrshortpl{RNA} containing 21 -- 28 \glsxtrlongpl{nt}}
\newglossaryentrywithacronym{ERC}{external RNA control}{external RNA control}{an approach for \textit{microarray performance assessment}; see \cite{Tong2006} for more details}

\newacronym{GO}{GO}{gene ontology}
\newacronym{DE}{DE}{differentially expressed}
\newacronym{GRN}{GRN}{gene regulatory network}
\newgacronymwithdesc{GRNN}{gene regulatory \glsentrylong{NN}}{gene regulatory\glsxtrlong{NN}}
\newgacronymwithdesc{QNN}{quadratic \glsentrylong{NN}}{quadratic \glsxtrlong{NN}}

\newacronym{ATAC-seq}{ATAC-seq}{Assay for Transposase-Accessible Chromatin using sequencing}
\newgacronymwithdesc{scATAC-Seq}{single-cell \glsentryshort{ATAC-seq}}{single-cell \glsxtrshort{ATAC-seq}}
\newacronym{snmC-Seq}{snmC-Seq}{single-nucleus methylcytosine sequencing}
\newacronym{scVI}{scVI}{single-cell variational inference}

\newglossaryentrywithacronym{GeNN}{genetic neural network}{genetic neural network}{a neural network architecture for \glsxtrlong{GE} tasks presented in \cite{Eetemadi2018}}
\newglossaryentrywithacronym{LinGeNN}{linear GeNN}{linear GeNN}{a variant of \glsxtrshort{GeNN} with linear activation function \cite{Eetemadi2018}}

\newglossaryentry{dgea_gls}{
    name={differential gene expression analysis},
    short={DGE analysis},
    long={differential gene expression analysis},
    description={a commonly used computational approach for identifying genes whose expressions are significantly different between two phenotypes \cite{Abbas2020}},
}

\newglossaryentry{dgea}{
    type=\acronymtype,
    name={DGE analysis},
    short={DGE analysis},
    long={differential gene expression analysis},
    description={differential gene expression analysis},
    first={differential gene expression (DGE) analysis\glsadd{dgea_gls}},
    plural={differential gene expression analyses},
    shortplural={DGE analyses},
    firstplural=differential gene expression (DGE) analyses\glsadd{dgea_gls},
    see=[Glossary:]{dgea_gls}
}

\newglossaryentry{DGEX}{
    name={D--GEX},
    description={\seeref{a \glsxtrlong{NN} for \glsxtrlong{GE} inference \cite{Chen2016b}}{sec:dgex}},
}
\newglossaryentry{dnamicroarray}{
    name={DNA microarray},
    description={used for measuring \glsxtrshort{DNA} levels \cite{Sealfon2010}; mainly for \glsxtrlong{GE} analysis},
}

\newglossaryentry{rnamicroarray}{
    name={RNA microarray},
    description={used for measuring \glsxtrshort{RNA} levels \cite{Sealfon2010}},
}
\newglossaryentry{microarray}{
    name={microarray},
    description={used for measuring, usually \glsxtrshort{DNA} or \glsxtrshort{RNA} levels \cite{Sealfon2010}; called \gls{dnamicroarray} when measuring \glsxtrshort{DNA} levels and \gls{rnamicroarray} when measuring \glsxtrshort{RNA} levels  \cite{Sealfon2010}; see \cite{Aparna2023} for other types},
}
\newglossaryentry{hybridization}{
    name={hybridization},
    description={a process in which a \glsxtrshort{cDNA} binds to the probes on the \gls{microarray} surface},
}

\newglossaryentry{L1000}{
    name={L1000},
    description={a cost-efficient \glsxtrlong{GE} assay, see \cite{Subramanian2017}},
}

\newglossaryentry{Hi-C}{
    name={Hi-C},
    description={high-throughput method for detection of chromatin interactions},
}

\newglossaryentry{k-means}{
    name={k-means},
    description={a greedy clustering algorithm},
}

\newglossaryentry{lasso}{
    name={lasso},
    description={a \glsxtrlong{LR} variant with $l_1$ regularization; more in \cite{Tibshirani1996}},
}

\newglossaryentrywithacronym{BP}{backpropagation}{backpropagation}{a method for calculation of the gradient of the loss function w.r.t. the network's weight}
\newglossaryentrywithacronym{SGD}{stochastic gradient descent}{stochastic gradient descent}{a method for optimization of an objective function; it is a variant of gradient descent that uses stochastic batches of data instead of the entire dataset to calculate the gradient}

\newglossaryentry{Adam}{
    name={Adam},
    description={a popular variant of the \glsxtrshort{SGD} optimization algorithm; more in \cite{Kingma2014}},
}

\newglossaryentrywithacronym{rRNA}{ribosomal RNA}{ribosomal RNA}{a primary component of ribosomes; non-coding \glsxtrshort{RNA}}
\newglossaryentrywithacronym{RNA-Seq}{RNA-Seq}{RNA sequencing}{ a \glsxtrshort{NGS} \glsxtrshort{RNA} sequencing method; allows for measuring \glsxtrlong{GE} levels}
\newgacronymwithdesc{scRNA-Seq}{single-cell \glsentryshort{RNA-Seq}}{single-cell \glsxtrshort{RNA-Seq}}
\newglossaryentrywithacronym{GEO}{Gene Expression Omnibus}{Gene Expression Omnibus}{a central and public repository of high-throughput \glsxtrlong{GE} data by \glsxtrshort{NCBI}, more in \cite{Edgar2002}}
\newglossaryentrywithacronym{PCA}{principal component analysis}{principal component analysis}{a dimensionality reduction method by linear transformation into a new coordinate system that respects the variance in the data}

\newglossaryentrywithacronym{RNN}{recurrent neural network}{recurrent neural network}{a type of \glsxtrlong{NN} that contains loop in the information flow}

\newglossaryentrywithacronym{BRNN}{bidirectional recurrent neural network}{bidirectional recurrent neural network}{a type of \glsxtrlong{RNN} where the outputs can use information from both past and future states, more in \cite{Schuster1997}}
\newacronym{gcd}{GCD}{Greatest Common Divisor}
\newglossaryentrywithacronymandref{ANN}{artificial neural network}{artificial neural network}{a biologically inspired computational model, interchangeably used with the term \glsxtrlong{NN}}{ch:background_nn}
\newglossaryentrywithacronymandref{NN}{neural network}{neural network}{a biologically inspired computational model, interchangeably used with the term \glsxtrlong{ANN}}{ch:background_nn}
\newglossaryentrywithacronym{GPU}{GPU}{graphics processing unit }{a specialized piece of hardware initially designed to accelerate image processing and computer graphics in general; often used for acceleration of training and inference of \glsxtrlongpl{NN}}

\newacronym{CPU}{CPU}{central processing unit}
\newacronym{LSTM}{LSTM}{long short-term memory}
\newacronym{BiLSTM}{BiLSTM}{bidirectional long short-term memory}
\newacronym{CNN}{CNN}{convolutional neural network}
\newacronym{GMDH}{GMDH}{group method of data handling}
\newacronym{KNN}{KNN}{k--nearest neighbor}
\newacronym{MNN}{MNN}{mutual nearest neighbor}
\newacronym[see=NN]{DKNN}{DKNN}{Deep Kronecker neural network}
\newacronym{FISH}{FISH}{fluorescence in situ hybridization}
\newacronym{nt}{nt}{nucleotide}
\newacronym{NGS}{NGS}{next-generation sequencing}
\newacronym{NCBI}{NCBI}{National Center for Biotechnology Information}
\newacronym{DL}{DL}{deep learning}
\newacronym[see=NN]{GNN}{GNN}{graph neural networks}
\newacronym{MR}{MR}{master regulator}
\newacronym{LP}{LP}{linear programming}
\newacronym{LR}{LR}{linear regression}
\newacronym{TRN}{TRN}{transcriptional regulatory network}
\newacronym{MLP}{MLP}{multi-layer perceptron}
\newacronym{NLP}{NLP}{natural language processing}
\newacronym{CDF}{CDF}{cumulative distribution function}
\newacronym{ML}{ML}{machine learning}
\newacronym{MP}{MP}{max-pooling}
\newacronym{ILSVRC}{ILSVRC}{ImageNet Large-Scale Visual Recognition Challenge}
\newacronym{NO}{NO}{neural operator}
\newacronym{NFT}{NFT}{non-fungible token}

\newglossaryentrywithacronym{SHAP}{Shapley additive explanations}{Shapley additive explanations}{a game theoretic approach for explaining the output of a \glsxtrlong{ML} model}

\newacronym{AF}{AF}{activation function}
\newglossaryentrywithacronym{AAF}{adaptive activation function}{adaptive \glsxtrlong{AF}}{an \glsxtrlong{AF} that can adapt to the data; often, it has a parameter whose value is data-dependent}
\newglossaryentrywithacronym{TAF}{trainable activation function}{trainable \glsxtrlong{AF}}{an \glsxtrlong{AF}; another name for the \glsxtrlong{AAF} (\glsxtrshort{AAF})}

\newglossaryentry{sigmoid}{
    name={sigmoid},
    long={sigmoid},
    description={a mathematical function having \textit{S-shaped} curve, \gls{logisticsigmoid} is the most known example},
}

\newglossaryentry{probit}{
    name={probit},
    description={\seeref{another name for the cumulative standard distribution function when used as \glsxtrlong{AF}}{sec:logistic_sigmoid}},
}
\newglossaryentry{logisticsigmoid}{
    name={logistic sigmoid},
    description={\seeref{one of most common \glsxtrlongpl{AF}  used in \glsxtrshortpl{NN}}{sec:logistic_sigmoid}},
}

\newglossaryentry{improvedlogisticsigmoid}{
    name={improved logistic sigmoid},
    description={\seeref{an \glsxtrlong{AF} based on the \gls{logisticsigmoid}}{sec:improved_logistic_sigmoid}},
}

\newglossaryentry{SigLin}{
    name={SigLin},
    description={\seeref{an \glsxtrlong{AF}; a combination of the \gls{logisticsigmoid} and a linear function}{sec:siglin}},
}

\newglossaryentry{scaledlogisticsigmoid}{
    name={scaled logistic sigmoid},
    description={\seeref{an \glsxtrlong{AF} based on the \gls{logisticsigmoid}}{sec:scaled_logistic_sigmoid}},
}
\newacronymwithrefdesc{LRTanh}{linearized \glsentrylong{tanh} unit}{linearized \glsxtrlong{tanh}}{sec:logistic_sigmoid}
\newglossaryentrywithacronymandref{tanh}{hyperbolic tangent}{hyperbolic tangent}{one of most common \glsxtrlongpl{AF}  used in \glsxtrshortpl{NN}}{sec:logistic_sigmoid}
\newacronymwithref{SSS}{shifted and scaled sigmoid}{sec:sss}
\newacronymwithrefdesc{VSF}{variant \gls{sigmoid} function}{variant \gls{sigmoid} function}{sec:vsf}
\newglossaryentrywithacronymandref{stanh}{scaled hyperbolic tangent}{scaled hyperbolic tangent}{a scaled variant of \glsxtrlong{tanh}}{sec:stanh}
\newacronymwithrefdesc{pRPPSG}{p-recursive piecewise polynomial sigmoid generator}{p-recursive piecewise polynomial \gls{sigmoid} generator}{sec:logistic_sigmoid}

\newglossaryentrywithacronymandref{arctan}{arctangent}{arctangent}{the inverse of the tangent function, also used as an \glsxtrlong{AF}}{sec:arctan}

\newglossaryentry{rectifiedhyperbolicsecant}{
    name={rectified hyperbolic secant},
    description={\seeref{an \glsxtrlong{AF} based on the hyperbolic secant}{sec:resech}},
}
\newacronymwithref{binary AF}{binary activation function}{sec:binaryaf}
\newacronymwithref{LiSHT}{linearly scaled hyperbolic tangent}{sec:lisht}
\newacronymwithref{PSGU}{Parameterized self-circulating gating unit}{sec:psgu}
\newglossaryentry{PATS}{
    name={PATS},
    description={\seeref{an \glsxtrlong{AF}; not an abbreviation}{sec:pats}},
}
\newacronymwithref{SiLU}{sigmoid-weighted linear unit}{sec:silu}
\newacronymwithref{WiG}{weighted sigmoid gate unit}{sec:silu}
\newacronymwithrefdesc{TS-swish}{triple-state \gls{swish} unit}{triple-state \gls{swish} unit}{sec:tsswish}
\newacronymwithrefdesc{TS-sigmoid}{triple-state \gls{sigmoid} unit}{triple-state \gls{sigmoid} unit}{sec:tssigmoid}
\newacronymwithref{GLU}{gated linear unit}{sec:glu}
\newacronymwithrefdesc{GTU}{gated \glsentryshort{tanh} unit}{gated \glsxtrshort{tanh} unit}{sec:gtu}
\newacronymwithrefdesc{ReGLU}{gated \glsentryshort{ReLU}}{gated \glsxtrshort{ReLU}}{sec:reglu}
\newacronymwithrefdesc{GEGLU}{gated \glsentryshort{GELU}}{gated \glsxtrshort{GELU}}{sec:geglu}
\newacronymwithrefdesc{SwiGLU}{gated \gls{swish}}{gated \gls{swish}}{sec:swiglu}
\newacronymwithrefdesc{dSiLU}{derivative of  \glsentrylong{SiLU}}{derivative of  \glsxtrlong{SiLU}}{sec:dsilu}
\newacronymwithrefdesc{DoubleSiLU}{double \glsentrylong{SiLU}}{double \glsxtrlong{SiLU}}{sec:double_silu}
\newacronymwithrefdesc{TSiLU}{\glsentrylong{tanh} \glsentrylong{SiLU}}{\glsxtrlong{tanh} \glsxtrlong{SiLU}}{sec:tsilu}
\newacronymwithrefdesc{ATSiLU}{arctan \glsentrylong{SiLU}}{arctan \glsxtrlong{SiLU}}{sec:atsilu}
\newglossaryentry{SwAT}{
    name={SwAT},
    description={\seeref{an \glsxtrlong{AF} combining \gls{logisticsigmoid} and arctan; not an abbreviation}{sec:swat}},
}
\newglossaryentry{E-swish}{
    name={E-swish},
    description={\seeref{an \glsxtrlong{AF} based on the \gls{swish}}{sec:eswish}},
}
\newacronymwithrefdesc{T-swish}{tunable \gls{swish}}{arctan \gls{swish}}{sec:tswish}
\newacronymwithrefdesc{RePSU}{rectified parametric \gls{sigmoid} unit}{rectified parametric \gls{sigmoid} unit}{sec:repsu}
\newacronymwithrefdesc{RePSKU}{rectified parametric \gls{sigmoid} shrinkage unit}{rectified parametric \gls{sigmoid} shrinkage unit}{sec:repsu}
\newacronymwithrefdesc{RePSHU}{rectified parametric \gls{sigmoid} stretchage unit}{rectified parametric \gls{sigmoid} stretchage unit}{sec:repsu}
\newacronymwithrefdesc{SSBS}{smooth \gls{sigmoid}-based shrinkage}{smooth \gls{sigmoid}-based shrinkage}{sec:repsu}
\newacronymwithref{AQuLU}{adaptive quadratic linear unit}{sec:aqulu}
\newacronymwithref{SinLU}{sinu-sigmoidal linear unit}{sec:sinlu}
\newglossaryentry{ErfAct}{
    name={ErfAct},
    description={\seeref{an \glsxtrlong{AAF} based on the \gls{swish}}{sec:erfact}},
}

\newacronymwithrefdesc{pserf}{parametric serf}{parametric \gls{serf}}{sec:pserf}
\newglossaryentry{swim}{
    name={swim},
    description={\seeref{an \glsxtrlong{AAF} similar to the \gls{swish}}{sec:swim}},
}

\newacronymwithref{MSiLU}{modified sigmoid-weighted linear unit}{sec:msilu}
\newacronymwithref{ptanh}{penalized hyperbolic tangent}{sec:penalized_hyperbolic_tangent}
\newacronymwithref{SLS-SS}{scaled logistic sigmoid with scaled sine}{sec:scaled_logistic_sigmoid}
\newacronymwithref{SRS}{soft-root-sign}{sec:srs}
\newacronymwithref{SC}{soft clipping}{sec:sc}
\newglossaryentry{hexpo}{
    name={Hexpo},
    description={\seeref{an \glsxtrlong{AF}}{sec:hexpo}},
}
\newglossaryentry{Elliott}{
    name={Elliott},
    description={\seeref{an \glsxtrlong{AF}}{sec:elliott}},
}
\newglossaryentrywithacronymandref{ReLU}{ReLU}{rectified linear unit}{one of the most popular \glsxtrlongpl{AF}}{sec:relu}

\newglossaryentry{softmax}{
    name={softmax},
    description={\seeref{a popular \glsxtrlong{AF} for classification problems; outputs a soft argmax of outputs of a given layer}{sec:softmax}},
}
\newacronymwithrefdesc{tsoftmax}{tuned softmax}{tuned \gls{softmax}}{sec:tsoftmax}
\newacronymwithrefdesc{glsoftmax}{generalized Lehmer softmax}{generalized Lehmer \gls{softmax}}{sec:glsoftmax}
\newacronymwithrefdesc{gpsoftmax}{generalized power softmax}{generalized power \gls{softmax}}{sec:gpsoftmax}
\newglossaryentry{betasoftmax}{
    name={\texorpdfstring{$\beta$}{Beta}-softmax},
    description={\seeref{an extension of the \gls{softmax} \gls{AF}}{sec:betasoftmax}},
}
\newglossaryentry{shifted_relu}{
    name={Shifted ReLU},
    description={\seeref{an \glsxtrlong{AF}; translated \glsxtrshort{ReLU}}{sec:shifted_relu}},
}
\newglossaryentrywithacronymandref{ShiLU}{ShiLU}{adaptive shifted ReLU}{a \glsxtrshort{ReLU} based \glsxtrlong{AF}; not to be confused with \gls{shifted_relu}}{sec:shilu}
\newglossaryentrywithacronymandref{LReLU}{leaky ReLU}{leaky ReLU}{a popular \glsxtrshort{ReLU} based \glsxtrlong{AF}; allows information flow for negative inputs unlike \glsxtrshort{ReLU}}{sec:lrelu}
\newglossaryentrywithacronymandref{RReLU}{randomized leaky ReLU}{randomized leaky ReLU}{a \glsxtrshort{LReLU} based \glsxtrlong{AF} with stochastic leakiness during training}{sec:rrelu}
\newglossaryentrywithacronymandref{S-RReLU}{softsign randomized leaky ReLU}{softsign randomized leaky ReLU}{a \glsxtrshort{RReLU} based \glsxtrlong{AF} combined with \gls{softsign}}{sec:srrelu}
\newacronymwithrefdesc{MarReLU}{margin \glsentryshort{PReLU}}{margin\glsxtrshort{PReLU}}{sec:marrelu}
\newacronymwithrefdesc{FunPReLU}{funnel \glsentrylong{PReLU}}{funnel \glsxtrlong{PReLU}}{sec:funprelu}
\newacronymwithrefdesc{FunReLU}{funnel \glsentrylong{ReLU}}{funnel \glsxtrlong{ReLU}}{sec:funprelu}
\newacronymwithrefdesc{RPReLU}{react-\glsentryshort{PReLU}}{react-\glsxtrshort{PReLU}}{sec:rprelu}

\newacronymwithref{SAU}{smooth activation unit}{sec:sau}
\newacronymwithref{SMU}{smooth maximum unit}{sec:smu}
\newglossaryentry{SMU1}{
    name={SMU-1},
    description={\seeref{an \glsxtrlong{AF}; a variant of the \glsxtrshort{SMU} using a different smoothing approach}{sec:lreluplus}},
}
\newacronymwithrefdesc{VLReLU}{very \glsentrylong{LReLU}}{very \glsxtrlong{LReLU}}{sec:lrelu}
\newacronymwithrefdesc{OLReLU}{optimized \glsentrylong{LReLU}}{optimized \glsxtrlong{LReLU}}{sec:lrelu}
\newacronymwithrefdesc{SlReLU}{Sloped ReLU}{Sloped \glsxtrshort{ReLU}}{sec:slrelu}
\newacronymwithrefdesc{NReLU}{noisy ReLU}{noisy \glsxtrshort{ReLU}}{sec:nrelu}
\newglossaryentry{SineReLU}{
    name={SineReLU},
    description={\seeref{an \glsxtrlong{AF}; extension of \glsxtrshort{ReLU}}{sec:sinerelu}},
}
\newglossaryentry{minsin}{
    name={minsin},
    description={\seeref{an \glsxtrlong{AF}}{sec:minsin}},
}
\newacronymwithref{VLU}{variational linear unit}{sec:vlu}
\newacronymwithrefdesc{PVLU}{parametric \glsentrylong{VLU}}{parametric \glsxtrlong{VLU}}{sec:pvlu}
\newacronymwithref{SCAA}{spatial context-aware activation}{sec:scaa}
\newacronymwithrefdesc{FReLU}{flexible ReLU}{flexible \glsxtrshort{ReLU}}{sec:frelu}
\newglossaryentry{StarReLU}{
    name={StarReLU},
    description={\seeref{an \glsxtrlong{AF}; extension of \glsxtrshort{ReLU}}{sec:starrelu}},
}
\newacronymwithrefdesc{AReLU}{Attention-based ReLU}{Attention-based \glsxtrshort{ReLU}}{sec:arelu}

\newacronymwithrefdesc{DPReLU}{dual parametric ReLU}{dual parametric \glsxtrshort{ReLU}}{sec:dprelu}
 \newglossaryentry{dual_line}{
    name={Dual Line},
    description={\seeref{an \glsxtrlong{AF}; extension of \glsxtrshort{DPReLU}}{sec:dual_line}},
}
\newacronymwithref{PiLU}{piecewise linear unit}{sec:pilu}
\newglossaryentrywithacronymandref{DPAF}{dual parametric activation function}{dual parametric activation function}{an \glsxtrlong{AF} inspired by \glsxtrshort{DPReLU}}{sec:dpaf}
\newglossaryentrywithacronymandref{FPAF}{fully parameterized activation function}{fully parameterized activation function}{an \glsxtrlong{AF} similar to \glsxtrshort{DPAF}}{sec:fpaf}
\newacronymwithrefdesc{EPReLU}{Elastic \glsentryshort{PReLU}}{Elastic \glsxtrshort{PReLU}}{sec:eprelu}
\newglossaryentry{RT-ReLU}{
    type=\acronymtype,
    name={RT--ReLU},
    short={RT--ReLU},
    long={randomly translational ReLU},
    first={randomly translational ReLU (RT--ReLU)},
    description={\seeref{randomly translational ReLU}{sec:rtrelu}},
}
\newacronymwithref{NLReLU}{natural-logarithm-ReLU}{sec:nlrelu}
\newacronymwithref{SLU}{softplus linear unit}{sec:slu}
\newacronymwithref{ReSP}{rectified softplus}{sec:resp}
\newacronymwithref{PReNU}{parametric rectified non-linear unit}{sec:prenu}
\newglossaryentrywithacronymandref{BReLU}{bounded ReLU}{bounded ReLU}{an \glsxtrlong{AF}; \glsxtrshort{ReLU} variant with bounds}{sec:brelu}
\newglossaryentry{hard_sigmoid}{
    name={Hard sigmoid},
    description={\seeref{an \glsxtrlong{AF} similar to \glsxtrshort{BReLU} with \gls{sigmoid}-like shape}{sec:hard_sigmoid}},
}
\newglossaryentry{hard_tanh}{
    name={HardTanh},
    description={\seeref{an \glsxtrlong{AF} similar to \gls{hard_sigmoid} with \glsxtrshort{tanh}-like shape}{sec:hard_tanh}},
}
\newglossaryentry{SvHardTanh}{
    type=\acronymtype,
    name={SvHardTanh},
    short={SvHardTanh},
    long={SvHardTanh},
    description={\seeref{\gls{hard_tanh} with a fixed vertical shift}{sec:shifted_hardtanh}},
}
\newglossaryentry{ShHardTanh}{
    type=\acronymtype,
    name={ShHardTanh},
    short={ShHardTanh},
    long={ShHardTanh},
    description={\seeref{\gls{hard_tanh} with a fixed horizontal shift}{sec:shifted_hardtanh}},
}
\newglossaryentry{hardshrink}{
    name={Hardshrink},
    description={\seeref{an \glsxtrlong{AF} similar to \gls{hard_sigmoid}}{sec:hardshrink}},
}
\newglossaryentry{softshrink}{
    name={Softshrink},
    description={\seeref{an \glsxtrlong{AF} similar to \gls{hard_sigmoid}}{sec:softshrink}},
}
\newacronymwithref{TRec}{truncated rectified}{sec:trec}
\newglossaryentry{hardswish}{
    name={Hard-Swish},
    description={\seeref{an \glsxtrlong{AF} similar to \gls{hard_sigmoid} related to the \gls{swish} function}{sec:hardswish}},
}
\newglossaryentrywithacronymandref{BLReLU}{bounded leaky ReLU}{bounded leaky ReLU}{an \glsxtrlong{AF}; \glsxtrshort{ReLU} variant combining \glsxtrshort{BReLU} and \glsxtrshort{LReLU}}{sec:blrelu}
\newacronymwithref{PLU}{piecewise linear unit}{sec:plu}
\newacronymwithref{AdaLU}{adaptive linear unit}{sec:adalu}
\newacronymwithrefdesc{TSAF}{trapezoid-shaped \glsentrylong{AF}}{trapezoid-shaped \glsxtrlong{AF}}{sec:tsaf}
\newacronymwithref{ARiA}{Adaptive Richard's curve weighted activation}{sec:aria}
\newglossaryentrywithacronymandref{ARiA2}{ARiA2}{Adaptive Richard's curve weighted activation 2}{an \glsxtrlong{AF}; special case of \glsxtrshort{ARiA}}{sec:aria}
\newacronymwithref{MTLU}{multi-bin trainable linear unit}{sec:mtlu}
\newacronymwithref{PWLU}{piecewise linear unit}{sec:mtlu}
\newacronymwithrefdesc{N-PWLU}{non-uniform \glsentrylong{PWLU}}{non-uniform \glsxtrlong{PWLU}}{sec:mtlu}
\newacronymwithrefdesc{CPN}{continuous piecewise nonlinear activation function}{continuous piecewise nonlinear \glsxtrlong{AF}}{sec:cpn}
\newacronymwithref{LuTU}{look-up table unit}{sec:lutu}
\newglossaryentry{maxout}{
    name={maxout unit},
    description={\seeref{an \glsxtrlong{AF} returning the maximum of several linear functions}{sec:maxout}},
}
\newacronymwithref{KAF}{kernel activation function}{sec:kaf}
\newacronym{SAVE}{SAVE}{structured activation of vertex entropy}
\newacronymwithref{vReLU}{V-shaped ReLU}{sec:vrelu}
\newglossaryentry{pan}{
    name={pan},
    description={\seeref{a piecewise linear \glsxtrlong{AF}}{sec:pan}},
}
\newacronymwithref{AbsLU}{absolute linear unit}{sec:abslu}
\newacronymwithrefdesc{mReLU}{mirrored \glsentrylong{ReLU}}{mirrored \glsxtrlong{ReLU}}{sec:mrelu}
\newacronymwithref{LRTLU}{leaky rectified triangle linear unit}{sec:lsptlu}
\newacronymwithref{LSPTLU}{leaky single-peaked triangle linear unit}{sec:lsptlu}
\newglossaryentry{tent}{
    name={tent},
    description={\seeref{an \glsxtrshort{ReLU}-based \glsxtrlong{AF}}{sec:tent}},
}
\newglossaryentry{hat}{
    name={hat},
    description={\seeref{an \glsxtrshort{ReLU}-based \glsxtrlong{AF}}{sec:hat}},
}
\newglossaryentry{softmodulusq}{
    name={SoftModulusQ},
    description={\seeref{an \glsxtrlong{AF}; a quadratic approximation of the \glsxtrshort{vReLU}}{sec:softmodulusq}},
}
\newglossaryentry{softmodulust}{
    name={SoftModulusT},
    description={\seeref{an \glsxtrlong{AF}; a \glsxtrshort{tanh} based approximation of the \glsxtrshort{vReLU}}{sec:softmodulust}},
}
\newglossaryentry{SignReLU}{
    name={SignReLU},
    description={\seeref{an \glsxtrlong{AF}; a combination of \glsxtrshort{ReLU} and \gls{softsign}}{sec:signrelu}},
}
\newglossaryentry{Li-ReLU}{
    name={Li-ReLU},
    description={\seeref{an \glsxtrlong{AF}; a combination of a linear function and \glsxtrshort{ReLU} }{sec:lirelu}},
}
\newglossaryentry{DLU}{
    name={DLU},
    description={\seeref{different name for \gls{SignReLU} used in \cite{Li2022ANewActivation,Pan2023Smoothing}}{sec:signrelu}},
}
\newacronymwithrefdesc{CReLU}{concatenated \glsentryshort{ReLU}}{concatenated \glsxtrshort{ReLU}}{sec:crelu}
\newacronymwithrefdesc{NCReLU}{negative \glsentryshort{CReLU}}{negative \glsxtrshort{CReLU}}{sec:ncrelu}
\newacronymwithrefdesc{BAF}{bipolar activation function}{bipolar \glsxtrlong{AF}}{sec:ncrelu}
\newglossaryentry{dualrelu}{
    name={DualReLU},
    description={\seeref{an \glsxtrlong{AF}; a two-dimensional \glsxtrshort{ReLU} variant}{sec:dualrelu}},
}
\newacronymwithref{OPLU}{orthogonal permutation liner unit}{sec:oplu}
\newglossaryentry{power_activation_function}{
    name={power activation function},
    description={\seeref{an \glsxtrlong{AF}; also known as \glsxtrshort{RePU}}{sec:repu}},
}

\newacronymwithrefdesc{EReLU}{elastic \glsentryshort{ReLU}}{elastic \glsxtrshort{ReLU}}{sec:erelu}
\newacronymwithref{RePU}{rectified power unit}{sec:repu}
\newacronymwithrefdesc{AppReLU}{approximate \glsentryshort{ReLU}}{approximate \glsxtrshort{ReLU}}{sec:apprelu}
\newacronymwithrefdesc{PLAF}{power linear \glsentrylong{AF}}{power linear \glsxtrlong{AF}}{sec:plaf}
\newacronymwithrefdesc{EPLAF}{even \glsentrylong{PLAF}}{even \glsxtrlong{PLAF}}{sec:plaf}
\newacronymwithrefdesc{OPLAF}{odd \glsentrylong{PLAF}}{odd \glsxtrlong{PLAF}}{sec:plaf}
\newacronymwithrefdesc{ABReLU}{average biased \glsentryshort{ReLU}}{average biased \glsxtrshort{ReLU}}{sec:abrelu}
\newacronymwithrefdesc{DRLU}{delay \glsentryshort{ReLU}}{delay \glsxtrshort{ReLU}}{sec:drlu}
\newacronymwithref{AOAF}{adaptive offset activation function}{sec:aoaf}
\newacronymwithref{DLReLU}{dynamic leaky ReLU}{sec:dlrelu}
\newacronymwithref{DReLU}{dynamic ReLU}{sec:drelu}
\newacronymwithref{DisReLU}{displaced ReLU}{sec:disrelu}
\newacronymwithrefdesc{MLReLU}{modified \glsentryshort{LReLU}}{modified \glsxtrshort{LReLU}}{sec:mlrelu}
\newacronymwithref{FTS}{flatted-T swish}{sec:fts}
\newglossaryentry{ReLUSwish}{
    name={ReLU-Swish},
    description={\seeref{an \glsxtrlong{AF}; special case of \glsxtrshort{FTS}}{sec:fts}},
}
\newacronymwithref{OAF}{Optimal Activation Functio}{sec:oaf}
\newglossaryentrywithacronymandref{ELU}{exponential linear unit}{exponential linear unit}{a popular \glsxtrlong{AF} extending \glsxtrshort{ReLU}}{sec:elu}
\newacronymwithref{REU}{rectified exponential unit}{sec:reu}
\newacronymwithref{ADA}{apical dendrite activation}{sec:ada}
\newacronymwithrefdesc{LADA}{leaky \glsentrylong{ADA}}{leaky \glsxtrlong{ADA}}{sec:lada}
\newacronymwithref{SigLU}{sigmoid linear unit}{sec:siglu}
\newacronymwithrefdesc{SaRa}{\gls{swish} and \glsentryshort{ReLU} activation}{\gls{swish} and \glsxtrshort{ReLU} activation}{sec:sara}
\newglossaryentry{maxsig}{
    name={maxsig},
    description={\seeref{an \glsxtrlong{AF}}{sec:maxsig}},
}
\newacronymwithrefdesc{ThLU}{\glsentryshort{tanh} linear unit}{\glsxtrshort{tanh} linear unit}{sec:thlu}
\newglossaryentry{maxtanh}{
    name={maxtanh},
    description={\seeref{an \glsxtrlong{AF}}{sec:thlu}},
}
\newglossaryentry{dualelu}{
    name={DualELU},
    description={\seeref{an \glsxtrlong{AF}; an \glsxtrshort{ELU} variant similar to \gls{dualrelu}}{sec:dualelu}},
}
\newacronymwithrefdesc{DiffELU}{difference \glsentrylong{ELU}}{difference \glsxtrlong{ELU}}{sec:diffelu}
\newacronymwithref{PolyLU}{polynomial linear unit}{sec:polylu}
\newglossaryentry{L-ReLU}{
    type=\acronymtype,
    name={L--ReLU},
    short={L--ReLU},
    long={Lipschitz ReLU},
    description={\seeref{Lipschitz \glsxtrshort{ReLU}}{sec:l-relu}},
}
\newacronymwithref{IpLU}{polynomial linear unit}{sec:iplu}
\newacronymwithref{PoLU}{power linear unit}{sec:polu}
\newacronymwithref{PFLU}{power function linear unit}{sec:pflu}
\newacronymwithrefdesc{FPFLU}{faster \glsentrylong{PFLU}}{faster \glsxtrlong{PFLU}}{sec:fpflu}
\newacronymwithref{EACU}{elastic adaptively parametric compounded unit}{sec:eacu}
\newacronymwithrefdesc{SELU}{scaled ELU}{scaled \glsxtrshort{ELU}}{sec:selu}
\newacronymwithref{SERLU}{scaled exponentially-regularized linear unit}{sec:serlu}
\newglossaryentry{ASERLU}{
    name={ASERLU},
    description={\seeref{an \glsxtrlong{AF}; an extension of the \glsxtrshort{SERLU} for \glsxtrshort{BiLSTM}  architectures}{sec:serlu}},
}
\newacronymwithref{LSELU}{leaky scaled exponential linear unit}{sec:lselu}
\newacronymwithref{sSELU}{scaled scaled exponential linear unit}{sec:sselu}
\newglossaryentry{rsigelu}{
    name={RSigELU},
    description={\seeref{an \glsxtrlong{AF}; a parametric \glsxtrshort{ELU}}{sec:rsigelu}},
}
\newglossaryentry{HardSReLUE}{
    name={HardSReLUE},
    description={\seeref{an \glsxtrlong{AF}; a parametric \glsxtrshort{ELU}}{sec:hardsrelue}},
}
\newacronymwithref{ELiSH}{exponential linear sigmoid squashing}{sec:elish}
\newacronymwithref{HardELiSH}{hard exponential linear sigmoid squashing}{sec:hardelish}
\newglossaryentry{rsigelud}{
    name={RSigELUD},
    description={\seeref{an \glsxtrlong{AF}; a variant of \gls{rsigelu} with two parameters}{sec:rsigelud}},
}
\newglossaryentry{LSReLU}{
    name={LS--ReLU},
    description={\seeref{an \glsxtrlong{AF} inspired by \gls{ReLU}; not an abbreviation}{sec:lsrelu}},
}
\newglossaryentry{SQNL}{
    name={SQNL},
    description={\seeref{an \glsxtrlong{AF}; not an abbreviation}{sec:sqnl}},
}
\newacronymwithref{SQLU}{square linear unit}{sec:sqlu}
\newacronymwithref{squish}{square swish}{sec:squish}
\newacronymwithref{SqSoftplus}{square softplus}{sec:sqsoftplus}
\newacronymwithref{LogSQNL}{square logistic sigmoid}{sec:logsqnl}
\newacronymwithref{SQMAX}{square softmax}{sec:sqmax}
\newacronymwithref{ISRLU}{inverse square root linear unit}{sec:isrlu}
\newacronymwithref{ISRU}{inverse square root unit}{sec:isru}
\newacronymwithrefdesc{MEF}{modified Elliott function}{modified \gls{Elliott} function}{sec:mef}
\newacronymwithref{LinQ}{linear quadratic activation}{sec:linq}
\newacronymwithrefdesc{SQRT}{square-root-based activation function}{square-root-based \glsxtrlong{AF}}{sec:sqrt}
\newglossaryentry{bent_identity}{
    name={bent identity},
    description={\seeref{an \glsxtrlong{AF}}{sec:bent_identity}},
}
\newacronymwithrefdesc{SSAF}{S-shaped activation function}{S-shaped \glsxtrlong{AF}}{sec:sqrt}
\newglossaryentry{Mishra}{
    name={Mishra},
    description={\seeref{an \glsxtrlong{AF}; unnamed in the original paper}{sec:mishra}},
}
\newacronymwithrefdesc{SBAF}{Saha-Bora activation function}{Saha-Bora \glsxtrlong{AF}}{sec:sbaf}
\newacronymwithrefdesc{LAF}{logarithmic activation function}{logarithmic \glsxtrlong{AF}}{sec:laf}
\newacronymwithref{SPOCU}{scaled polynomial constant unit}{sec:spocu}
\newacronymwithrefdesc{PUAF}{polynomial \glsentrylong{UAF}}{polynomial \glsxtrlong{UAF}}{sec:puaf}
\newglossaryentry{softplus}{
    name={softplus},
    description={\seeref{an \glsxtrlong{AF}}{sec:softplus}},
}
\newacronymwithref{PSoftplus}{parametric softplus}{sec:psoftplus}
\newacronymwithref{RSP}{rand softplus}{sec:rsp}
\newglossaryentry{softsign}{
    name={softsign},
    description={\seeref{an \glsxtrlong{AF}}{sec:softsign}},
}
\newglossaryentry{smooth_step}{
    name={smooth step},
    description={\seeref{an \glsxtrlong{AF}}{sec:smooth_step}},
}
\newglossaryentry{mish}{
    name={mish},
    description={\seeref{an \glsxtrlong{AF}; combination of \glsxtrshort{tanh} and \gls{softplus}}{sec:mish}},
}
\newglossaryentry{smish}{
    name={smish},
    description={\seeref{an \glsxtrlong{AF}; combination of \glsxtrshort{tanh}, logarithm, and \gls{logisticsigmoid}}{sec:smish}},
}
\newacronymwithrefdesc{SC-mish}{soft clipping mish}{soft clipping \gls{mish}}{sec:scmish}
\newacronymwithrefdesc{SC-swish}{soft clipping swish}{soft clipping \gls{swish}}{sec:scswish}
\newacronymwithrefdesc{p-swish}{parametric swish}{parametric\gls{swish}}{sec:pswish}
\newacronymwithrefdesc{SCL-mish}{soft clipping learnable mish}{soft clipping learnable \gls{mish}}{sec:scmish}
\newglossaryentry{TanhExp}{
    name={TanhExp},
    description={\seeref{an \glsxtrlong{AF}; combination of \glsxtrshort{tanh} and exponential function}{sec:mish}},
}
\newglossaryentry{serf}{
    name={serf},
    description={\seeref{an \glsxtrlong{AF}; combination of the Gauss error function and \gls{softplus}}{sec:serf}},
}
\newacronymwithref{EANAF}{efficient asymmetric nonlinear activation function}{sec:eanaf}
\newglossaryentry{sinsig}{
    name={SinSig},
    description={\seeref{an \glsxtrlong{AF}; uses \gls{logisticsigmoid} and is similar to \gls{mish} and \gls{swish}}{sec:sinsig}},
}
\newacronymwithrefdesc{SiELU}{\glsentrylong{GELU} with \gls{sigmoid} \glsentrylong{AF}}{\glsxtrlong{GELU} with \gls{sigmoid} \glsxtrlong{AF}}{sec:sielu}
\newglossaryentry{Aranda-Ordaz}{
    name={Aranda-Ordaz},
    description={\seeref{an \glsxtrlong{AF}}{sec:arandaordaz}},
}
\newacronymwithref{bfire}{bi-firing activation function}{sec:bfire}
\newacronymwithref{bbfire}{bounded bi-firing activation function}{sec:bbfire}
\newacronymwithrefdesc{PMAF}{piecewise Mexican-hat activation function}{piecewise Mexican-hat \glsxtrlong{AF}}{sec:pmaf}
\newacronymwithrefdesc{PRBF}{piecewise \glsentrylong{RBF}}{piecewise \glsxtrlong{RBF}}{sec:prbf}
\newglossaryentrywithacronymandref{GELU}{GELU}{Gaussian error linear unit}{a popular \glsxtrlong{AF} based on the cumulative distribution function of the normal distribution}{sec:gelu}
\newacronymwithref{SGELU}{symmetrical Gaussian error linear unit}{sec:sgelu}
\newacronymwithref{CaLU}{Cauchy linear unit}{sec:calu}
\newacronymwithref{LaLU}{Laplace linear unit}{sec:lalu}
\newacronymwithref{CoLU}{Collapsing linear unit}{sec:colu}
\newacronymwithref{ASSF}{adaptive slope sigmoidal function}{sec:assf}
\newacronymwithref{SG}{Sigmoid-Gumbel}{sec:sg}
\newglossaryentry{comb-H-sine}{
    name={comb-H-sine},
    description={\seeref{an \glsxtrlong{AF}}{sec:combhsine}},
}
\newacronymwithref{m-arcsinh}{modified arcsinh}{sec:marcsinh}
\newglossaryentry{hyper-sinh}{
    name={hyper-sinh},
    description={\seeref{an \glsxtrlong{AF}}{sec:hypersinh}},
}

\newglossaryentry{arctid}{
    name={arctid},
    description={\seeref{an \glsxtrlong{AF}}{sec:arctid}},
}
\newglossaryentry{sinc}{
    name={sinc},
    description={\seeref{an \glsxtrlong{AF}}{sec:sinc}},
}
\newglossaryentry{polyexp}{
    name={polyexp},
    description={\seeref{an \glsxtrlong{AF}}{sec:polyexp}},
}
\newglossaryentry{E-Tanh}{
    name={E-Tanh},
    description={\seeref{an \glsxtrlong{AF}}{sec:etanh}},
}
\newglossaryentry{wave}{
    name={wave},
    description={\seeref{an \glsxtrlong{AF}}{sec:wave}},
}
\newglossaryentry{cosid}{
    name={cosid},
    description={\seeref{an \glsxtrlong{AF}}{sec:cosid}},
}
\newglossaryentry{sinp}{
    name={sinp},
    description={\seeref{an \glsxtrlong{AF} based on the sine function}{sec:sinp}},
}
\newacronymwithref{GCU}{growing cosine unit}{sec:gcu}
\newacronymwithref{ASU}{amplifying sine unit}{sec:asu}
\newacronymwithref{SSU}{shifted sine unit}{sec:sinc}
\newacronymwithref{DSU}{decaying sine unit}{sec:dsu}
\newacronymwithref{NCU}{non-monotonic cubic unit}{sec:ncu}
\newglossaryentry{triple}{
    name={triple},
    description={\seeref{an \glsxtrlong{AAF}}{sec:triple}},
}
\newacronymwithref{SQU}{shifted quadratic unit}{sec:squ}
\newacronymwithref{HcLSH}{hyperbolic cosine linearized squashing function}{sec:hclsh}
\newacronymwithref{KDAC}{knowledge discovery activation function}{sec:kdac}
\newacronym{WTA}{WTA}{winner-take-all}
\newglossaryentrywithacronymandref{k-WTA}{k-\glsentrylong{WTA}}{k-\glsentrylong{WTA}}{an \glsxtrlong{AF} based on the \glsentrylong{WTA} principle}{sec:kwta}
\newacronymwithrefdesc{VBAF}{volatility-based activation function}{volatility-based  \glsxtrlong{AF}}{sec:vbaf}
\newacronymwithrefdesc{CAF}{chaotic activation function}{chaotic \glsxtrlong{AF}}{sec:caf}
\newacronymwithrefdesc{HCAF}{hybrid chaotic activation function}{hybrid \glsxtrlong{CAF}}{sec:hcaf}
\newacronymwithrefdesc{FCAF}{fusion of chaotic activation function}{fusion of \glsxtrlong{CAF}}{sec:fcaf}
\newacronymwithrefdesc{CCAF}{cascade chaotic activation function}{cascade \glsxtrlong{CAF}}{sec:ccaf}
\newglossaryentry{sincos}{
    name={sincos},
    description={\seeref{an \glsxtrlong{AF}}{sec:sincos}},
}
\newacronymwithref{CSS}{combination of sine and logistic sigmoid}{sec:css}
\newacronymwithrefdesc{CatAF}{catalytic activation function}{catalytic \glsxtrlong{AF}}{sec:cataf}
\newglossaryentry{expcos}{
    name={expcos},
    description={\seeref{an \glsxtrlong{AF}}{sec:expcos}},
}
\newglossaryentry{rootsig}{
    name={rootsig},
    description={\seeref{an \glsxtrlong{AF}}{sec:rootsig}},
}
\newglossaryentry{nsigmoid}{
    name={n-sigmoid},
    description={\seeref{a \gls{sigmoid}-based \glsxtrlong{AF}}{sec:logistic_sigmoid}},
}
\newglossaryentrywithacronymandref{PReLU}{PReLU}{parametric rectified linear unit}{a popular \glsxtrlong{AAF}; a \glsxtrshort{LReLU} variant with trainable leakiness}{sec:prelu}
\newglossaryentry{PPReLU}{
    type=\acronymtype,
    name={positive PReLU},
    short={PReLU\textsuperscript{+}},
    long={positive PReLU},
    description={\seeref{positive \gls{PReLU}}{sec:pprelu}},
}
\newacronymwithrefdesc{LeLeLU}{leaky learnable \glsentryshort{ReLU}}{leaky learnable \glsxtrshort{ReLU}}{sec:lelelu}
\newacronymwithref{PREU}{parametric rectified exponential unit}{sec:preu}
\newglossaryentry{RT-PReLU}{
    type=\acronymtype,
    name={RT--PReLU},
    short={RT--PReLU},
    long={randomly translational PReLU},
    description={\seeref{randomly translational \glsxtrshort{PReLU}}{sec:rtprelu}},
    first={randomly translational PReLU (RT--PReLU)}
}
\newglossaryentry{ProbAct}{
    type=\acronymtype,
    name={ProbAct},
    short={ProbAct},
    long={ProbAct},
    description={\seeref{probabilistic activation}{sec:probact}},
}
\newglossaryentry{paired_relu}{
    name={paired ReLU},
    description={\seeref{paired \glsxtrshort{ReLU}}{sec:paired_relu}},
}
\newacronymwithrefdesc{RMAF}{ReLU memristor-like activation function}{\glsxtrshort{ReLU} memristor-like \glsxtrlong{AF}}{sec:rmaf}

\newacronymwithrefdesc{PTELU}{parametric tanh linear unit}{parametric \glsxtrshort{tanh} linear unit}{sec:ptelu}
\newacronymwithrefdesc{TaLU}{tanh linear unit}{\glsxtrshort{tanh} linear unit}{sec:talu}
\newglossaryentry{PTaLU}{
    name={PTaLU},
    description={\seeref{an \glsxtrshort{AAF}, \glsxtrshort{TaLU} variant with another parameter}{sec:ptalu}},
}\newglossaryentry{tanhLU}{
    name={tanhLU},
    description={\seeref{an \glsxtrshort{AAF}, combination of \glsxtrshort{tanh} and a linear function}{sec:tanhlu}},
}
\newacronymwithrefdesc{TReLU}{tanh based ReLU}{\glsxtrshort{tanh} based \glsxtrshort{ReLU}}{sec:trelu}
\newacronymwithrefdesc{TReLU2}{TReLU variant 2}{\glsxtrshort{TReLU} variant 2}{sec:trelu}
\newacronymwithrefdesc{TeLU}{tanh \glsentrylong{ELU}}{\glsxtrshort{tanh} \glsxtrlong{ELU}}{sec:telu}
\newacronymwithrefdesc{ReLTanh}{rectified linear tanh}{rectified linear \glsxtrshort{tanh}}{sec:reltanh}
\newacronymwithrefdesc{PELU}{parametric \glsentrylong{ELU}}{parametric \glsxtrlong{ELU}}{sec:pelu}
\newacronymwithrefdesc{FELU}{fast \glsentrylong{ELU}}{fast \glsxtrshort{ELU}}{sec:felu}
\newglossaryentry{PFELU}{
    name={P+FELU},
    description={\seeref{trainable \glsxtrshort{FELU} based \glsxtrshort{AF}}{sec:pfelu}},
}
\newacronymwithrefdesc{MPELU}{multiple \glsentrylong{PELU}}{multiple \glsxtrlong{PELU}}{sec:mpelu}
\newglossaryentry{P-E2-X}{
    name={P-E2-XU},
    description={\seeref{family of \glsxtrlongpl{AAF}}{sec:pe2relu}},
}
\newglossaryentry{P-E2-ReLU}{
    name={P-E2-ReLU},
    description={\seeref{an \glsxtrlong{AAF} combining two \glsxtrshortpl{ELU} and \glsxtrshort{ReLU}}{sec:pe2relu}},
}
\newglossaryentry{P-E2-Id}{
    name={P-E2-Id},
    description={\seeref{an \glsxtrlong{AAF} based on \gls{P-E2-ReLU}}{sec:pe2relu}},
}
\newglossaryentry{P-E2-ReLU-1}{
    name={P-E2-ReLU-1},
    description={\seeref{an \glsxtrlong{AAF} based on \gls{P-E2-ReLU}}{sec:pe2relu}},
}
\newacronymwithref{BLU}{bendable linear unit}{sec:blu}
\newacronymwithrefdesc{ReBLU}{rectified \glsentryshort{BLU}}{rectified \glsxtrshort{BLU}}{sec:reblu}
\newglossaryentry{DELU}{
    name={DELU},
    description={\seeref{an \glsxtrlong{AF} proposed in \cite{Pishchik2023}; not an abbreviation}{sec:delu}},
}
\newglossaryentry{soft_exponential}{
    name={soft exponential},
    description={\seeref{an \glsxtrlong{AF} interpolating between logarithmic, linear, and exponential functions}{sec:soft_exponential}},
}
\newacronymwithrefdesc{CELU}{continuously differentiable \glsentrylong{ELU}}{continuously differentiable \glsxtrlong{ELU}}{sec:celu}
\newacronymwithrefdesc{ErfReLU}{Erf-based \glsentryshort{ReLU}}{Erf-based \glsxtrshort{ReLU}}{sec:erfrelu}
\newacronymwithrefdesc{PSELU}{parametric scaled \glsentrylong{ELU}}{parametric scaled \glsxtrlong{ELU}}{sec:pselu}
\newacronymwithrefdesc{LPSELU}{leaky \glsentrylong{PSELU}}{leaky \glsxtrlong{PSELU}}{sec:lpselu}
\newglossaryentry{LPSELURP}{
    type=\acronymtype,
    name={LPSELU\_RP},
    short={LPSELU\_RP},
    long={\glsentrylong{LPSELU} with reposition parameter},
    description={\seeref{\glsxtrlong{LPSELU} with reposition parameter}{sec:lpselurp}},
}
\newglossaryentry{shifted_elu}{
    name={shifted \glsentrylong{ELU}},
    short={shifted \glsentryshort{ELU}},
    long={shifted \glsentrylong{ELU}},
    shortplural={shifted \glsentryshort{ELU}s},
    longplural={shifted \glsentrylong{ELU}s},
    description={\seeref{an \glsxtrlong{AF}; an \glsxtrshort{ELU} with a vertical shift (\glsxtrshort{SvELU}, \glsxtrshort{PSvELU}) or an \glsxtrshort{ELU} with  ahorizontal shift (\glsxtrshort{ShELU}, \glsxtrshort{PShELU})}{sec:shifted_elus}},
}
\newglossaryentry{SvELU}{
    type=\acronymtype,
    name={SvELU},
    short={SvELU},
    long={SvELU},
    description={\seeref{\glsxtrlong{ELU} with a fixed vertical shift}{sec:shifted_elus}},
}
\newglossaryentry{ShELU}{
    type=\acronymtype,
    name={ShELU},
    short={ShELU},
    long={ShELU},
    description={\seeref{\glsxtrlong{ELU} with a fixed horizontal shift}{sec:shifted_elus}},
}
\newglossaryentry{PSvELU}{
    type=\acronymtype,
    name={PSvELU},
    short={PSvELU},
    long={PSvELU},
    description={\seeref{\glsxtrlong{ELU} with a trainable vertical shift}{sec:shifted_elus}},
}
\newglossaryentry{PShELU}{
    type=\acronymtype,
    name={PShELU},
    short={PShELU},
    long={PShELU},
    description={\seeref{\glsxtrlong{ELU} with a trainable horizontal shift}{sec:shifted_elus}},
}
\newacronymwithrefdesc{PDELU}{parametric deformable \glsentrylong{ELU}}{parametric deformable \glsxtrlong{ELU}}{sec:pdelu}
\newacronymwithrefdesc{EDELU}{extended \glsentrylong{ELU}}{extended \glsxtrlong{ELU}}{sec:edelu}
\newacronymwithrefdesc{EELU}{elastic \glsentrylong{ELU}}{elastic \glsxtrlong{ELU}}{sec:eelu}
\newacronymwithref{PFPLUS}{parametric first power linear unit with sign}{sec:pfplus}
\newacronymwithref{FPLUS}{first power linear unit with sign}{sec:polylu}
\newglossaryentry{generalized hyperbolic tangent}{
    name={generalized hyperbolic tangent},
    description={\seeref{an \glsxtrlong{AAF}}{sec:generalized_hyperbolic_tangent}},
}
\newacronymwithrefdesc{SVAF}{slope varying \glsentrylong{AF}}{slope varying \glsxtrlong{AF}}{sec:svaf}
\newglossaryentry{TanhSoft}{
    name={TanhSoft},
    description={\seeref{a family of \glsxtrlongpl{AAF} proposed in \cite{Biswas2021TanhSoft}}{sec:tanhsoft}},
}
\newacronymwithrefdesc{psigmoid}{parametric sigmoid}{parametric \gls{sigmoid}}{sec:psigmoid}
\newglossaryentry{swish}{
    name={swish},
    description={\seeref{an \glsxtrlong{AAF}; an adaptive variant of \glsxtrshort{SiLU}}{sec:swish}},
}
\newacronymwithrefdesc{AHAF}{adaptive hybrid \glsentrylong{AF}}{adaptive hybrid \glsxtrlong{AF}}{sec:ahaf}
\newglossaryentrywithacronymandref{PSiLU}{PSiLU}{parametric SiLU}{another name for the \gls{swish} \glsxtrshort{AF}}{sec:swish}
\newacronymwithrefdesc{PSSiLU}{parametric shifted \glsentryshort{SiLU}}{parametric shifted \glsxtrshort{SiLU}}{sec:pssilu}
\newacronymwithrefdesc{ACON}{activate or not \glsentrylong{AF}}{activate or not \glsxtrlong{AF}}{sec:swish}
\newglossaryentry{ACON-A}{
    name={ACON-A},
    description={\seeref{an \glsxtrlong{AAF} from the \glsxtrshort{ACON} family; another name for the \gls{swish}}{sec:swish}},
}
\newglossaryentry{ACON-B}{
    name={ACON-B},
    description={\seeref{an \glsxtrlong{AAF} from the \glsxtrshort{ACON} family; extension of the \gls{ACON-A}}{sec:aconb}},
}
\newglossaryentry{ACON-C}{
    name={ACON-C},
    description={\seeref{an \glsxtrlong{AAF} from the \glsxtrshort{ACON} family; extension of the \gls{ACON-B}}{sec:aconc}},
}
\newglossaryentry{MetaACON}{
    name={MetaACON},
    description={\seeref{an extension of the \glsxtrshort{ACON} family where the parameter $a_i$ is determined by a small \glsxtrshort{NN}}{sec:aconc}},
}
\newglossaryentry{MetaACON-C}{
    name={MetaACON-C},
    description={\seeref{an \gls{ACON-C} variant where the parameter $a_i$ is determined by a small \glsxtrshort{NN}}{sec:aconc}},
}
\newglossaryentry{1Dmeta-ACON}{
    name={1Dmeta-ACON},
    description={\seeref{a \gls{MetaACON} variant}{sec:aconc}},
}
\newglossaryentry{gish}{
    name={gish},
    description={\seeref{an \glsxtrlong{AF} based on the \glsxtrshort{SiLU}}{sec:gish}},
}
\newglossaryentry{logish}{
    name={logish},
    description={\seeref{an \glsxtrlong{AF} based on the \glsxtrshort{SiLU}}{sec:logish}},
}
\newglossaryentry{LogLogish}{
    name={LogLogish},
    description={\seeref{an \glsxtrlong{AF} based on the \glsxtrshort{SiLU} and the \gls{LogLog}}{sec:loglogish}},
}
\newglossaryentry{ExpExpish}{
    name={ExpExpish},
    description={\seeref{an \glsxtrlong{AF} based on the \glsxtrshort{SiLU}}{sec:expexpish}},
}
\newglossaryentry{selfarctan}{
    name={self arctan},
    description={\seeref{an \glsxtrlong{AF} based on the \glsxtrshort{SiLU}}{sec:selfarctan}},
}
\newglossaryentry{phish}{
    name={phish},
    description={\seeref{an \glsxtrlong{AF} based on the \glsxtrshort{SiLU} and \glsxtrshort{GELU}}{sec:phish}},
}
\newglossaryentry{suish}{
    name={suish},
    description={\seeref{an \glsxtrlong{AF}; proposed as the alternative to the \glsxtrshort{SiLU} and \gls{swish}}{sec:suish}},
}
\newacronymwithrefdesc{TSReLU}{tangent-sigmoid ReLU}{\glslink{tanh}{tangent}-\gls{sigmoid} \glsxtrshort{ReLU}}{sec:tsrelu}
\newacronymwithrefdesc{TBSReLU}{tangent-bipolar-sigmoid ReLU}{\glslink{tanh}{tangent}-bipolar-\gls{sigmoid} \glsxtrshort{ReLU}}{sec:tbsrelu}
\newacronymwithrefdesc{TSReLUl}{TSReLU learnable}{\glsxtrshort{TSReLU} learnable}{sec:psgu}
\newacronymwithrefdesc{TBSReLUl}{TBSReLU learnable}{\glsxtrshort{TBSReLU} learnable}{sec:tbsrelul}
\newacronymwithrefdesc{pLogish}{parametric logish}{parametric \gls{logish}}{sec:plogish}
\newacronymwithrefdesc{ARBF}{adaptive \glsentrylong{RBF}}{parametric \glsxtrlong{RBF}}{sec:arbf}
\newacronymwithrefdesc{PGELU}{parametric \glsentrylong{GELU}}{parametric \glsxtrlong{GELU}}{sec:pgelu}
\newacronymwithrefdesc{PSF}{parametric \glsentrylong{sigmoid} function}{parametric \glsxtrlong{sigmoid} function}{sec:psf}
\newacronymwithrefdesc{STAC-tanh}{slope and threshold \glsentrylong{AAF} function with \glsentryshort{tanh} function}{slope and threshold \glsxtrlong{AAF} function with \glsxtrshort{tanh} function}{sec:stactanh}
\newacronymwithref{GRA}{generalized Riccati activation}{sec:gra}
\newacronymwithrefdesc{PFTS}{parametric \glsentrylong{FTS}}{parametric \glsxtrlong{FTS}}{sec:pfts}
\newacronymwithrefdesc{PFPM}{parametric flatten-p \gls{mish}}{parametric flatten-p \gls{mish}}{sec:pfpm}
\newacronymwithref{GEU}{Gaussian error unit}{sec:geu}
\newacronymwithrefdesc{SGT}{scaled-gamma-\glsentryshort{tanh}}{scaled-gamma-\glsxtrshort{tanh}}{sec:sgt}
\newacronymwithref{RSign}{react-sign}{sec:rsign}
\newglossaryentry{P-SIG-RAMP}{
    name={P-SIG-RAMP},
    description={\seeref{an \glsxtrlong{AAF} combining \gls{logisticsigmoid} and \glsxtrshort{ReLU}}{sec:psigramp}},
}
\newacronymwithrefdesc{PSTanh}{parametric scaled \glsentrylong{tanh}}{parametric scaled \glsxtrlong{tanh}}{sec:pstanh}
\newacronymwithrefdesc{SSinH}{scaled sine-hyperbolic function}{scaled sine-hyperbolic function}{sec:ssinh}
\newacronymwithrefdesc{SExp}{scaled exponential function}{scaled exponential function}{sec:sexp}
\newacronymwithrefdesc{MWF}{modified Weibull function}{modified Weibull function}{sec:mwf}
\newacronymwithref{LAU}{logmoid activation unit}{sec:lau}
\newglossaryentry{symlog}{
    name={symlog},
    description={\seeref{an alternative name of the \glsxtrshort{LAU}}{sec:lau}},
}
\newglossaryentry{symexp}{
    name={symexp},
    description={\seeref{an \glsxtrlong{AF} inverse of the \glsxtrshort{LAU}}{sec:symexp}},
}
\newacronymwithref{CosLU}{cosinu-sigmoidal linear unit}{sec:coslu}
\newacronymwithref{AGumb}{adaptive Gumbel}{sec:AGumb}
\newglossaryentry{NewSigmoid}{
    name={NewSigmoid},
    description={\seeref{an \glsxtrlong{AF} similar to \gls{logisticsigmoid}}{sec:newsigmoid}},
}
\newglossaryentry{generalized_swish}{
    name={generalized swish},
    description={\seeref{an \glsxtrlong{AF} related to the \gls{SiLU}}{sec:generalized_swish}},
}
\newglossaryentry{exponential_swish}{
    name={exponential swish},
    description={\seeref{an \glsxtrlong{AF} related to the \gls{SiLU}}{sec:exponential_swish}},
}
\newglossaryentry{Sigmoid-Algebraic}{
    name={Sigmoid-Algebraic},
    description={\seeref{an \glsxtrlong{AF} similar to \gls{logisticsigmoid}}{sec:sigmoidalgebraic}},
}
\newglossaryentry{SincSigmoid}{
    name={Sinc-Sigmoid},
    description={\seeref{an \glsxtrlong{AF}}{sec:sincsigmoid}},
}
\newglossaryentry{root2sigmoid}{
    name={root2sigmoid},
    description={\seeref{an \glsxtrlong{AF} similar to \gls{logisticsigmoid}}{sec:root2sigmoid}},
}
\newglossaryentry{LogLog}{
    name={LogLog},
    description={\seeref{an \glsxtrlong{AF}}{sec:loglog}},
}
\newacronymwithrefdesc{cLogLog}{complementary LogLog}{complementary \gls{LogLog}}{sec:cloglog}
\newacronymwithrefdesc{cLogLogm}{modified \glsentryshort{cLogLog}}{modified \glsxtrshort{cLogLog}}{sec:cloglog}
\newglossaryentry{SechSig}{
    name={SechSig},
    description={\seeref{an \glsxtrlong{AF}}{sec:sechsig}},
}
\newacronymwithrefdesc{pSechSig}{parametric SechSig}{parametric \gls{SechSig}}{sec:sechsig}
\newglossaryentry{TanhSig}{
    name={TanhSig},
    description={\seeref{an \glsxtrlong{AF}}{sec:tanhsig}},
}
\newacronymwithrefdesc{pTanhSig}{parametric TanhSig}{parametric \gls{TanhSig}}{sec:tanhsig}
\newacronymwithrefdesc{MSAF}{multistate \glsentrylong{AF}}{multistate \glsxtrlong{AF}}{sec:msaf}
\newacronymwithrefdesc{SymMSAF}{symmetrical \glsentryshort{MSAF}}{symmetrical \glsxtrshort{MSAF}}{sec:msaf}
\newglossaryentry{scaled_softsign}{
    name={scaled softsign},
    description={\seeref{an \glsxtrlong{AAF}; an adaptive variant of \glsxtrlong{softsign}}{sec:scaled_softsign}},
}
\newglossaryentry{parameterized_softplus}{
    type=\acronymtype,
    name={parameterized softplus},
    short={s\textsubscript{+}2L},
    long={parameterized softplus},
    description={\seeref{parametrized \gls{softplus}}{sec:parameterized_softplus}},
}
\newacronymwithrefdesc{UAF}{universal \glsentrylong{AF}}{universal \glsxtrlong{AF}}{sec:uaf}
\newacronymwithrefdesc{LEAF}{learnable extended \glsentrylong{AF}}{learnable extended \glsxtrlong{AF}}{sec:leaf}
\newacronymwithrefdesc{GReLU}{generalized \glsentryshort{ReLU}}{generalized \glsxtrshort{ReLU}}{sec:grelu}
\newacronymwithrefdesc{MAF}{multiquadratic \glsentrylong{AF}}{multiquadratic \glsxtrlong{AF}}{sec:maf}
\newglossaryentry{EIS}{
    name={EIS},
    description={\seeref{a family of \glsxtrlongpl{AF}, not an abbreviation}{sec:eis}},
}
\newacronymwithref{GLN}{global-local neuron}{sec:gln}
\newacronymwithrefdesc{NAF}{neuron-adaptive \glsentrylong{AF}}{neuron-adaptive \glsxtrlong{AF}}{sec:naf}
\newacronymwithrefdesc{LAAF}{locally \glsentrylong{AAF}}{locally \glsxtrlong{AAF}}{sec:laaf}
\newacronymwithrefdesc{GAAF}{globally \glsentrylong{AAF}}{globally \glsxtrlong{AAF}}{sec:laaf}
\newacronymwithrefdesc{SAAAF}{shape autotuning \glsentrylong{AAF}}{shape autotuning \glsxtrlong{AAF}}{sec:saaaf}
\newacronymwithrefdesc{FAAF}{fractional \glsentrylong{AAF}}{fractional \glsxtrlong{AAF}}{sec:faaf}
\newacronymwithrefdesc{FracReLU}{fractional ReLU}{fractional \glsxtrshort{ReLU}}{sec:fracrelu}
\newacronymwithrefdesc{FracSoftplus}{fractional softplus}{fractional \gls{softplus}}{sec:fracsoftplus}
\newacronymwithrefdesc{FracTanh}{fractional tanh}{fractional \glsxtrshort{tanh}}{sec:fractanh}
\newacronymwithrefdesc{FracLReLU}{fractional LReLU}{fractional \glsxtrshort{LReLU}}{sec:fraclrelu}
\newacronymwithrefdesc{FracPReLU}{fractional PReLU}{fractional \glsxtrshort{PReLU}}{sec:fracprelu}
\newacronymwithrefdesc{FracELU}{fractional ELU}{fractional \glsxtrshort{ELU}}{sec:fracelu}
\newacronymwithrefdesc{FracSiLU}{fractional SiLU}{fractional \glsxtrshort{SiLU}}{sec:fracsilu}
\newacronymwithrefdesc{FracSiLU1}{FracSiLU variant 1}{\glsxtrshort{FracSiLU} variant 1}{sec:fracsilu}
\newacronymwithrefdesc{FracSiLU2}{FracSiLU variant 1}{\glsxtrshort{FracSiLU} variant 2}{sec:fracsilu}
\newacronymwithrefdesc{FracGELU}{fractional GELU}{fractional \glsxtrshort{GELU}}{sec:fracgelu}
\newacronymwithrefdesc{FracGELU1}{FracGELU variant 1}{\glsxtrshort{FracGELU} variant 1}{sec:fracgelu}
\newacronymwithrefdesc{FracGELU2}{FracGELU variant 1}{\glsxtrshort{FracGELU} variant 2}{sec:fracgelu}
\newacronymwithref{FALU}{fractional adaptive linear unit}{sec:falu}
\newacronymwithref{APLU}{adaptive piece-wise linear unit}{sec:aplu}
\newacronymwithrefdesc{SAAF}{smooth \glsentrylong{AAF}}{smooth \glsxtrlong{AAF}}{sec:aplu}
\newacronymwithref{SPLASH}{simple piecewise linear and adaptive function with symmetric hinges}{sec:splash}
\newacronymwithref{MBA}{multi-bias activation}{sec:mba}
\newacronymwithrefdesc{MeLU}{Mexican \glsentryshort{ReLU}}{Mexican \glsxtrshort{ReLU}}{sec:melu}
\newacronymwithrefdesc{MMeLU}{modified \glsentrylong{MeLU}}{modified \glsxtrlong{MeLU}}{sec:mmelu}
\newacronymwithrefdesc{GaLU}{Gaussian \glsentryshort{ReLU}}{Gaussian \glsxtrshort{ReLU}}{sec:galu}
\newacronymwithrefdesc{SReLU}{S-shaped \glsentryshort{ReLU}}{S-shaped \glsxtrshort{ReLU}}{sec:srelu}
\newglossaryentry{N-activation}{
    name={N-activation},
    description={\seeref{an \glsxtrlong{AAF} resembling the letter N}{sec:nactivation}},
}
\newacronymwithref{LiSA}{linearized sigmoidal activation}{sec:alisa}
\newacronymwithrefdesc{ALiSA}{adaptive \glsentryshort{LiSA}}{adaptive \glsxtrlong{LiSA}}{sec:alisa}
\newacronymwithrefdesc{All-ReLU}{alternated left \glsentryshort{ReLU}}{alternated left \glsxtrlong{ReLU}}{sec:allrelu}
\newacronymwithref{ABU}{adaptive blending unit}{sec:abu}
\newacronymwithrefdesc{TCA}{trained compound activation function}{trained compound \glsxtrlong{AF}}{sec:tca}
\newacronymwithrefdesc{TCAv2}{\glsxtrlong{TCA} variant 2}{\glsxtrlong{TCA} variant 2}{sec:tca}
\newacronymwithrefdesc{APAF}{average of a pool of activation functions}{average of a pool of \glsxtrlongpl{AF}}{sec:apaf}
\newacronymwithrefdesc{GABU}{gating \glsentrylong{ABU}}{gating \glsxtrlong{ABU}}{sec:gabu}

\newacronymwithrefdesc{SLAF}{self-learnable \glsentrylong{AF}}{self-learnable \glsxtrlong{AF}}{sec:slaf}
\newacronymwithrefdesc{ChPAF}{Chebyshev polynomial-based \glsentrylong{AF}}{Chebyshev polynomial-based \glsxtrlong{AF}}{sec:chpaf}
\newacronymwithrefdesc{LPAF}{Legendre polynomial-based \glsentrylong{AF}}{Legendre polynomial-based \glsxtrlong{AF}}{sec:lpaf}
\newacronymwithrefdesc{HPAF}{Hermite polynomial-based \glsentrylong{AF}}{Hermite polynomial-based \glsxtrlong{AF}}{sec:hpaf}
\newacronymwithref{MoGU}{mixture of Gaussian unit}{sec:mogu}
\newacronymwithref{FSA}{Fourier series activation}{sec:fsa}
\newacronymwithref{PAU}{Padé activation unit}{sec:pau}
\newacronymwithrefdesc{RPAU}{randomized \glsentrylong{PAU}}{randomized \glsxtrlong{PAU}}{sec:rpau}
\newacronymwithrefdesc{OPAU}{orthogonal \glsentrylong{PAU}}{orthogonal \glsxtrlong{PAU}}{sec:opau}
\newacronymwithref{ERA}{enhanced rational activation}{sec:era}
\newacronymwithrefdesc{SAF}{spline interpolating \glsentrylong{AF}}{spline interpolating \glsxtrlong{AF}}{sec:saf}
\newacronymwithrefdesc{PPAF}{piecewise polynomial \glsentrylong{AF}}{piecewise polynomial \glsxtrlong{AF}}{sec:saf}
\newacronymwithref{TruG}{truncated gaussian unit}{sec:trug}
\newacronymwithref{MSRF}{mollified square root function}{sec:msrf}
\newglossaryentry{SquarePlus}{
    name={SquarePlus},
    description={\seeref{an \glsxtrlong{AF} proposed in \cite{Barron2021}}{sec:squareplus}},
}
\newglossaryentry{StepPlus}{
    name={StepPlus},
    description={\seeref{an \glsxtrlong{AF} proposed in \cite{Pan2023Smoothing}}{sec:stepplus}},
}
\newglossaryentry{BipolarPlus}{
    name={BipolarPlus},
    description={\seeref{an \glsxtrlong{AF} proposed in \cite{Pan2023Smoothing}}{sec:stepplus}},
}
\newglossaryentry{LReLUPlus}{
    name={LReLUPlus},
    description={\seeref{an \glsxtrlong{AF}; a smoothed variant of \glsxtrshort{LReLU}}{sec:lreluplus}},
}
\newglossaryentry{vReLUPlus}{
    name={vReLUPlus},
    description={\seeref{an \glsxtrlong{AF}; a smoothed variant of \glsxtrshort{vReLU}}{sec:vreluplus}},
}
\newglossaryentry{SoftshrinkPlus}{
    name={SoftshrinkPlus},
    description={\seeref{an \glsxtrlong{AF}; a smoothed variant of \gls{softshrink}}{sec:softshrinkplus}},
}
\newglossaryentry{PanPlus}{
    name={PanPlus},
    description={\seeref{an \glsxtrlong{AF}; a smoothed variant of \gls{pan} function}{sec:panplus}},
}
\newglossaryentry{BReLUPlus}{
    name={BReLUPlus},
    description={\seeref{an \glsxtrlong{AF}; a smoothed variant of \glsxtrshort{BReLU}}{sec:breluplus}},
}
\newglossaryentry{SReLUPlus}{
    name={SReLUPlus},
    description={\seeref{an \glsxtrlong{AF}; a smoothed variant of \glsxtrshort{SReLU}}{sec:sreluplus}},
}
\newglossaryentry{HardTanhPlus}{
    name={HardTanhPlus},
    description={\seeref{an \glsxtrlong{AF}; a smoothed variant of \gls{hard_tanh} function}{sec:hardtanhplus}},
}
\newglossaryentry{HardshrinkPlus}{
    name={HardshrinkPlus},
    description={\seeref{an \glsxtrlong{AF}; a smoothed variant of \gls{hardshrink}}{sec:hardshrinkplus}},
}
\newglossaryentry{MeLUPlus}{
    name={MeLUPlus},
    description={\seeref{an \glsxtrlong{AF}; a smoothed variant of \glsxtrshort{MeLU}}{sec:meluplus}},
}
\newglossaryentry{TSAFPlus}{
    name={TSAFPlus},
    description={\seeref{an \glsxtrlong{AF}; a smoothed variant of \glsxtrshort{TSAF}}{sec:tsafplus}},
}
\newglossaryentry{ELUPlus}{
    name={ELUPlus},
    description={\seeref{an \glsxtrlong{AF}; a mollified variant of \glsxtrshort{ELU}}{sec:eluplus}},
}
\newglossaryentry{SwishPlus}{
    name={SwishPlus},
    description={\seeref{an \glsxtrlong{AF}; a mollified variant of \gls{swish}}{sec:swishplus}},
}
\newglossaryentry{MishPlus}{
    name={MishPlus},
    description={\seeref{an \glsxtrlong{AF}; a mollified variant of \gls{mish}}{sec:mishplus}},
}
\newglossaryentry{LogishPlus}{
    name={LogishPlus},
    description={\seeref{an \glsxtrlong{AF}; a mollified variant of \gls{logish}}{sec:logishplus}},
}
\newglossaryentry{SoftsignPlus}{
    name={SoftsignPlus},
    description={\seeref{an \glsxtrlong{AF}; a mollified variant of \gls{softsign}}{sec:softsignplus}},
}
\newglossaryentry{SignReLUPlus}{
    name={SignReLUPlus},
    description={\seeref{an \glsxtrlong{AF}; a smoothed variant of \glsxtrshort{SignReLU}}{sec:signreluplus}},
}
\newacronym{NIN}{NIN}{network in network}
\newacronym{MIN}{MIN}{maxout-in-network}
\newacronym{WHE}{WHE}{wide hidden expansion}

\newglossaryentry{NPF}{
    name={NPF},
    description={\seeref{an \glsxtrlong{AAF} proposed in \cite{Eisenach2017}; based on Nonparametric Fourier
    Basis Expansion}{sec:complex_approaches}},
}
\newacronymwithrefdesc{VAF}{variable \glsentrylong{AF}}{variable \glsxtrlong{AF}}{sec:vaf}
\newacronymwithref{FAB}{flexible activation bag}{sec:fab}
\newacronymwithrefdesc{DY--ReLU}{dynamic parameter \glsentryshort{ReLU}}{dynamic parameter \glsxtrshort{ReLU}}{sec:dyrelu}

\newacronym{MCDNN}{MCDNN}{multi-column deep neural network}
\newacronym{PC}{PC}{Parallel Circuit}
\newacronym{SNN}{SNN}{self-normalizing neural network}

\newacronym{RVFLN}{RVFLN}{random vector functional link network}
\newgacronymwithdesc{pRVFLN}{parsimonious \glsentrylong{RVFLN}}{parsimonious \glsxtrlong{RVFLN}}
\newgacronymwithdesc{M-RVFLN}{M-estimation-based \glsentryshort{RVFLN}}{M-estimation-based \glsxtrshort{RVFLN}}
\newacronymwithcustomshortdesc{epsRVFLN}{\ce{\epsilon}-HRVFLN}{\glsentryshort{RVFLN} with \ce{\epsilon}-insensitive Huber loss function}{\glsxtrshort{RVFLN} with \ce{\epsilon}-insensitive Huber loss function}
\newgacronymwithdesc{MK-RVFLN}{multi-kernel \glsentryshort{RVFLN}}{multi-kernel \glsxtrshort{RVFLN}}
\newgacronymwithdesc{K-RVFLN}{kernel \glsentryshort{RVFLN}}{kernel \glsxtrshort{RVFLN}}
\newgacronymwithdesc{EVWCA-MKRVFLN}{\glsentryshort{MK-RVFLN} with evaporation-based water cycle based parameter optimization}{\glsxtrshort{MK-RVFLN} with evaporation-based water cycle based parameter optimization}
\newgacronymwithdesc{WCRVFLN}{wavelet-coupled \glsentryshort{RVFLN}}{wavelet-coupled \glsxtrshort{RVFLN}}
\newgacronymwithdesc{SP-RVFLN}{sparse pre-trained \glsentryshort{RVFLN}}{sparse pre-trained \glsxtrshort{RVFLN}}
\newgacronymwithdesc{CRVFLN}{convolutional \glsentryshort{RVFLN}}{convolutional \glsxtrshort{RVFLN}}

\newacronym{SCN}{SCN}{stochastic configuration network}

\newgacronymwithdesc{LPSCN}{locality preserving \glsentryshort{SCN}}{locality preserving \glsxtrshort{SCN}}
\newgacronymwithdesc{RS-SCN}{\glsentryshort{SCN} with rough set based attribute reduction}{\glsxtrshort{SCN} with rough set based attribute reduction}
\newgacronymwithdesc{GA-SCN}{\glsentryshort{SCN} based on genetic algorithms}{\glsxtrshort{SCN} based on genetic algorithms}
\newgacronymwithdesc{G-BAPSO-SCN}{\glsentryshort{SCN} with hybrid bat-particle swarm optimization}{\glsxtrshort{SCN} with hybrid bat-particle swarm optimization}
\newgacronymwithdesc{MoGL-SCN}{Bayesian robust \glsentryshort{SCN} based on a mixture of the Gaussian and Laplace distributions}{Bayesian robust \glsxtrshort{SCN} based on a mixture of the Gaussian and Laplace distributions}
\newgacronymwithdesc{OSCN}{orthogonal \glsentryshort{SCN}}{orthogonal \glsxtrshort{SCN}}
\newgacronymwithdesc{FSCN}{fast \glsentryshort{SCN}}{fast \glsxtrshort{SCN}}
\newgacronymwithdesc{ISSA-FSCN}{\glsentryshort{FSCN} with an improved sparrow search algorithm}{\glsxtrshort{FSCN} with an improved sparrow search algorithm}
\newgacronymwithdesc{BSCN}{bidirectional \glsentryshort{SCN}}{bidirectional \glsxtrshort{SCN}}
\newgacronymwithdesc{CSSA-SCN}{chaotic sparrow search algorithm based \glsentryshort{SCN}}{chaotic sparrow search algorithm based \glsxtrshort{SCN}}
\newgacronymwithdesc{DSCN}{deep \glsentryshort{SCN}}{deep \glsxtrshort{SCN}}
\newgacronymwithdesc{PRSCN}{pruning regularization \glsentryshort{SCN}}{pruning regularization \glsxtrshort{SCN}}
\newgacronymwithdesc{SCBNN}{stochastic configured Bayesian \glsentrylong{NN}}{stochastic configured Bayesian \glsxtrlong{NN}}

\newacronym{ELM}{ELM}{extreme learning machine}
\newgacronymwithdesc{DAELM}{domain adaptation \glsentryshort{ELM}}{domain adaptation \glsxtrshort{ELM}}
\newgacronymwithdesc{ML-OCELM}{multilayer \glsentrylong{NN} based one-class classification with \glsentryshort{ELM}}{multilayer \glsxtrlong{NN} based one-class classification with \glsxtrshort{ELM}}
\newgacronymwithdesc{Fuzzy-ELM}{fuzzy \glsentryshort{ELM}}{fuzzy \glsxtrshort{ELM}}
\newgacronymwithdesc{ASELM}{daptive semi-supervised \glsentryshort{ELM}}{daptive semi-supervised \glsxtrshort{ELM}}
\newgacronymwithdesc{K-ELM}{kernel based \glsentryshort{ELM}}{kernel based \glsxtrshort{ELM}}
\newgacronymwithdesc{OS-ELM}{online sequential \glsentryshort{ELM}}{online sequential \glsxtrshort{ELM}}
\newgacronymwithdesc{SAO-ELM}{structure-adjustable \glsentryshort{OS-ELM}}{structure-adjustable \glsxtrshort{OS-ELM}}
\newgacronymwithdesc{DOS-ELM}{dynamic forgetting factor based \glsentryshort{OS-ELM} algorithm}{dynamic forgetting factor based \glsxtrshort{OS-ELM} algorithm}
\newgacronymwithdesc{FOS-ELM}{fuzziness-based \glsentryshort{OS-ELM} algorithm}{fuzziness-based \glsxtrshort{OS-ELM} algorithm}
\newgacronymwithdesc{ML-DOS-ELM}{multilayer \glsentryshort{DOS-ELM}}{\glsxtrshort{DOS-ELM}}
\newgacronymwithdesc{I-ELM}{incremental \glsentryshort{ELM}}{incremental \glsxtrshort{ELM}}
\newgacronymwithdesc{CI-ELM}{convex \glsentrylong{I-ELM}}{convex \glsxtrlong{I-ELM}}
\newgacronymwithdesc{cWOB-ELM}{coiflet wavelet-based optimization method-based \glsentryshort{ELM}}{coiflet wavelet-based optimization method-based \glsxtrshort{ELM}}
\newgacronymwithdesc{ML-ELM}{multi-layer \glsentrylong{ELM}}{multi-layer \glsxtrlong{ELM}}
\newgacronymwithdesc{H-ELM}{hierarchical \glsentrylong{ELM}}{hierarchical \glsxtrlong{ELM}}
\newgacronymwithdesc{D-HELM}{densely connected \glsentryshort{D-HELM}}{densely connected \glsxtrshort{D-HELM}}
\newgacronymwithdesc{ES-ELM}{evolutionary optimized \glsentryshort{ELM}}{evolutionary optimized \glsxtrshort{ELM}}
\newgacronymwithdesc{EM-ELM}{error minimized \glsentrylong{ELM}}{error minimized \glsxtrlong{ELM}}
\newgacronymwithdesc{BELM}{bayesian \glsentrylong{ELM}}{bayesian \glsxtrlong{ELM}}
\newgacronymwithdesc{ADHKELM}{adaptive deep hybrid kernel  \glsentrylong{ELM}}{adaptive deep hybrid kernel  \glsxtrlong{ELM}}

\newacronym{SVM}{SVM}{support vector machine}
\newacronym{RC}{RC}{reservoir computing}
\newacronym{ESN}{ESN}{echo state network}
\newgacronymwithdesc{PESN}{polynomial \glsentryshort{ESN}}{polynomial \glsxtrshort{ESN}}
\newgacronymwithdesc{DeepESN}{deep \glsentrylong{ESN}}{deep \glsxtrlong{ESN}}
\newgacronymwithdesc{S-PESN}{simplified \glsentryshort{PESN}}{simplified \glsxtrshort{PESN}}
\newgacronymwithdesc{VML-ESN}{variable memory length \glsentryshort{ESN}}{variable memory length \glsxtrshort{ESN}}
\newgacronymwithdesc{DRESN}{double-reservoir \glsentryshort{ESN}}{double-reservoir \glsxtrshort{ESN}}
\newgacronymwithdesc{MI-ESN}{mutual information optimized \glsentryshort{ESN}}{mutual information optimized \glsxtrshort{ESN}}
\newgacronymwithdesc{DBEN}{deep belief \glsentrylong{ESN}}{deep belief \glsxtrlong{ESN}}
\newgacronymwithdesc{MR-ESN}{multiple reservoirs \glsentrylong{ESN}}{multiple reservoirs \glsxtrlong{ESN}}
\newgacronymwithdesc{ESN-DE}{differential evolution based \glsentryshort{ESN}}{differential evolution based \glsxtrshort{ESN}}
\newgacronymwithdesc{O-ESN}{particle swarm optimized \glsentryshort{ESN}}{particle swarm optimized \glsxtrshort{ESN}}
\newacronymwithcustomshortdesc{piESN}{\ce{\pi}-ESN}{probabilistic \glsentryshort{ESN}}{probabilistic \glsxtrshort{ESN}}
\newgacronymwithdesc{HCR-ESN}{hybrid circle reservoir \glsentryshort{ESN}}{hybrid circle reservoir \glsxtrshort{ESN}}
\newgacronymwithdesc{SESN}{sinusoidal \glsentryshort{ESN}}{sinusoidal \glsxtrshort{ESN}}
\newgacronymwithdesc{FSDESN}{fast subspace decomposition \glsentrylong{ESN}}{fast subspace decomposition \glsxtrlong{ESN}}
\newgacronymwithdesc{RESN}{robust \glsentrylong{ESN}}{robust \glsxtrlong{ESN}}
\newgacronymwithdesc{FESN}{functional \glsentryshort{ESN}}{functional \glsxtrshort{ESN}}
\newgacronymwithdesc{TWIESN}{time warp invariant \glsentrylong{ESN}}{time warp invariant \glsxtrlong{ESN}}
\newacronym{SVESM}{SVESM}{support vector echo-state vector machine}
\newgacronymwithdesc{ESGNN}{echo state \glsentrylong{GNN}}{echo state \glsxtrlong{GNN}}
\newgacronymwithdesc{NGRC}{next generation \glsentrylong{RC}}{next generation \glsxtrlong{RC}}

\newacronym{RCN}{RCN}{random convolution node}

\newgacronymwithdesc{DNN}{deep \glsentrylong{NN}}{deep \glsxtrlong{NN}}
\newgacronymwithdesc{FFNN}{feed-forward \glsentrylong{NN}}{feed-forward \glsxtrlong{NN}}
\newacronym{GCN-RW}{GCN-RW}{graph convolutional networks with random weights}
\newacronym{RMDL}{RMDL}{random multimodel deep learning}
\newacronym{STM}{STM}{short-term memory}
\newacronym{GAN}{GAN}{generative adversarial network}
\newacronym{WRN}{WRN}{wide residual network}
\newacronym{RBM}{RBM}{restricted Boltzmann machine}
\newacronym{DBN}{DBN}{deep belief network}
\newacronym{DBM}{DBM}{deep Boltzmann machine}
\newacronym{RBF}{RBF}{radial basis function}
\newgacronymwithdesc{MI-CDBN}{mode isolation \glsentrylong{CDBN}}{mode isolation \glsxtrlong{CDBN}}
\newgacronymwithdesc{CDBN}{convolutional \glsentrylong{DBN}}{convolutional \glsxtrlong{DBN}}
\newgacronymwithdesc{GARBM}{Gaussian \glsentryshort{RBM} with binary auxiliary units}{Gaussian \glsxtrshort{RBM} with binary auxiliary units}
\newacronym{EBM}{EBM}{energy-based model}
\newacronym{AE}{AE}{autoencoder}
\newgacronymwithdesc{DAE}{denoising \glsentrylong{AE}}{denoising \glsxtrlong{AE}}
\newgacronymwithdesc{VAE}{variational \glsentrylong{AE}}{variational \glsxtrlong{AE}}
\newgacronymwithdesc{SAE}{sparse \glsentrylong{AE}}{sparse \glsxtrlong{AE}}
\newgacronymwithdesc{AVAE}{autoencoder \glsentryshort{VAE}}{autoencoder \glsxtrshort{VAE}}
\newgacronymwithdesc{PuVAE}{purifying \glsentryshort{VAE}}{purifying \glsxtrshort{VAE}}
\newacronym{TSP}{TSP}{travelling salesman problem}
\newgacronymwithdesc{EGAN}{edge adversarial  \glsentryshort{GAN}}{edge adversarial  \glsxtrshort{GAN}}
\newgacronymwithdesc{WGAN-GP}{Wasserstein \glsentryshort{GAN} with gradient penalty}{Wasserstein \glsxtrshort{GAN} with gradient penalty}
\newgacronymwithdesc{AC-GAN}{auxiliary classifier \glsentryshort{GAN}}{auxiliary classifier \glsxtrshort{GAN}}
\newgacronymwithdesc{TransGAN}{transformer based \glsentryshort{GAN}}{transformer based \glsxtrshort{GAN}}
\newgacronymwithdesc{SAGAN}{self-attention \glsentryshort{GAN}}{self-attention \glsxtrshort{GAN}}
\newgacronymwithdesc{STGAN}{selective transfer \glsentryshort{GAN}}{selective transfer \glsxtrshort{GAN}}
\newgacronymwithdesc{GCD-GAN}{gradient-guided dual-branch \glsentryshort{GAN}}{gradient-guided dual-branch \glsxtrshort{GAN}}
\newgacronymwithdesc{RI-GAN}{\glsentryshort{GAN} with residual inception modules}{ \glsxtrshort{GAN} with residual inception modules}
\newgacronymwithdesc{AutoGAN}{\glsentryshort{GAN} with neural architecture search}{\glsxtrshort{GAN} with neural architecture search}
\newgacronymwithdesc{PLGAN}{panoptic layout \glsentryshort{GAN}}{panoptic layout \glsxtrshort{GAN}}
\newgacronymwithdesc{DF-GAN}{deep fusion \glsentryshort{GAN}}{deep fusion \glsxtrshort{GAN}}
\newgacronymwithdesc{CMAFGAN}{cross-modal attention gusion based \glsentryshort{GAN}}{cross-modal attention gusion based \glsxtrshort{GAN}}
\newgacronymwithdesc{GGAN}{graph \glsentryshort{GAN}}{graph \glsxtrshort{GAN}}
\newgacronymwithdesc{AM-GAN}{fused \glsentryshort{GAN} with attention mechanism}{fused \glsxtrshort{GAN} with attention mechanism}
\newgacronymwithdesc{DGattGAN}{dual Generator attentional \glsentryshort{GAN}}{dual Generator attentional \glsxtrshort{GAN}}
\newgacronymwithdesc{CML-GAN}{contrastive meta-learning \glsentryshort{GAN}}{contrastive meta-learning \glsxtrshort{GAN}}
\newgacronymwithdesc{D2GAN}{dual discriminator \glsentryshort{GAN}}{dual discriminator \glsxtrshort{GAN}}
\newgacronymwithdesc{D2WMGAN}{dual discriminator weighted mixture \glsentryshort{GAN}}{dual discriminator weighted mixture \glsxtrshort{GAN}}
\newgacronymwithdesc{RoCGAN}{robust conditional \glsentryshort{GAN}}{robust conditional \glsxtrshort{GAN}}
\newgacronymwithdesc{VARGAN}{variance enforcing \glsentryshort{GAN}}{variance enforcing \glsxtrshort{GAN}}
\newgacronymwithdesc{DPGAN}{dual-stream \glsentryshort{GAN} with phase awareness}{dual-stream \glsxtrshort{GAN} with phase awareness}
\newgacronymwithdesc{GAGAN}{geometry-aware \glsentryshort{GAN}}{geometry-aware \glsxtrshort{GAN}}
\newgacronymwithdesc{RePGAN}{\glsentryshort{GAN} with residual partial modules}{\glsxtrshort{GAN} with residual partial modules}
\newgacronymwithdesc{SE-DCGAN}{squeeze-excitation network-deep convolution \glsentryshort{GAN}}{squeeze-excitation network-deep convolution \glsxtrshort{GAN}}
\newgacronymwithdesc{EA-GAN}{example attention \glsentryshort{GAN}}{example attention \glsxtrshort{GAN}}
\newgacronymwithdesc{HGAN}{hyperbolic \glsentryshort{GAN}}{hyperbolic \glsxtrshort{GAN}}
\newgacronymwithdesc{CRRAGAN}{cascading residual--residual attention \glsentryshort{GAN}}{cascading residual--residual attention \glsxtrshort{GAN}}
\newgacronymwithdesc{CycleGAN}{cycle-consistent \glsentryshort{GAN}}{cycle-consistent \glsxtrshort{GAN}}
\newgacronymwithdesc{cscGAN}{conditional single-cell \glsentryshort{GAN}}{conditional single-cell \glsxtrshort{GAN}}

\newgacronymwithdesc{MRI}{magnetic resonance imaging}{magnetic resonance imaging}
\newgacronymwithdesc{sMRI}{structural \glsentryshort{MRI}}{structural \glsxtrshort{MRI}}
\newacronym{CT}{CT}{computed tomography}
\newacronym{SPECT}{SPECT}{single-photon emission computed tomography}
\newacronym{PPGN}{PPGN}{Plug and Play generative networks}
\newacronym{DM}{DM}{diffusion model}
\newacronym{DDPM}{DDPM}{denoising diffusion probabilistic model}
\newacronym{SCIBER}{SCIBER}{single-cell integrator and batch effect remover}

\newacronym{DP}{DP}{dynamic programming}

\newglossaryentrywithacronymandref{TAAF}{transformative adaptive activation function}{transformative adaptive activation function}{a class of \glsxtrlongpl{AAF} allowing for translation and scaling of an activation function; proposed in this work}{sec:taaf}
\newglossaryentrywithacronymandref{ATU}{adaptive transformative unit}{adaptive transformative unit}{a proposed unit, used for the implementation of the proposed \glsxtrshort{TAAF}}{sec:atu}
\newgacronymwithdesc{DTAAF}{dual \glsentrylong{TAAF}}{dual \glsxtrlong{TAAF}}
\newgacronymwithdesc{GDTAAF}{generalized \glsentrylong{DTAAF}}{generalized \glsxtrlong{DTAAF}}

\newacronym{MAE}{MAE}{mean absolute error}
\newacronym{RMS}{RMS}{root mean square}
\newacronym{MSE}{MSE}{mean squared error}
\newacronym{SSE}{SSE}{sum of squared errors}
\newacronym{CI}{CI}{confidence interval}
\newacronym{MCC}{MCC}{Matthew's correlation coefficient}
\newgacronymwithdesc{MMAE}{mean \glsentrylong{MAE}}{mean \glsxtrlong{MAE}}
\newacronymwithref{MDAE}{mean difference of absolute errors}{eq:mdae}
\newacronymwithrefdesc{MMDAE}{mean \glsentryshort{MDAE}}{mean \glsxtrshort{MDAE}}{eq:mmdae}

\newacronym{FPGA}{FPGA}{field-programmable gate array}

\makeglossaries

\endgroup

%% file: section_motivation.tex
\section{Activations as special cases of TAAFs}
\label{sec:taaf_special_cases}
As already mentioned above, the \gls{TAAF} generalizes several other \glsxtrlongpl{AF} --- while the individual parameters were often proposed individually in the literature, the \gls{TAAF} provides a unique combination achieving better performance than if only some subset of parameters was used (see \cite{Kunc2024Decades} \cref{sec:params_importance} for experimental results).

The scaled hyperbolic tangent \cite{Lecun1998} (see \cite{Kunc2024Decades} \cref{sec:stanh}) can be considered as a special case of nonadaptive variant of \gls{TAAF} if the \gls{TAAF} is parametrized as $\alpha = a$, $\beta = b$, $\gamma = 0$, $\delta = 0$ and $f(z) = \tanh(z)$. Another case of nonadaptive \gls{TAAF} is the \gls{E-Tanh} (see \cite{Kunc2024Decades} \cref{sec:etanh}) that uses a fixed parameter $a$ for vertical scaling of the function; the \gls{TAAF} equivalent is, therefore, $\alpha = a$, $\beta = 1$, $\gamma = 0$, $\delta = 0$ and $f(z) = \exp(z)\tanh(z)$. 

The \gls{SSS} (see \cite{Kunc2024Decades} \cref{sec:sss}) is also a special case of a nonadaptive \gls{TAAF} as it is only  a \gls{logisticsigmoid} with horizontal scaling and translation; the \gls{TAAF} equivalent is therefore $\alpha = 1$, $\beta = a$, $\gamma = -ab$, $\delta = 0$ and $f(z) = \sigma(z)$. Similarly, the \gls{VSF} (see \cite{Kunc2024Decades} \cref{sec:vsf}) is also a translated and scaled \gls{logisticsigmoid}; its nonadaptive \gls{TAAF} equivalent is $\alpha = a$, $\beta = b$, $\gamma = 0$, $\delta = -c$ and $f(z) = \sigma(z)$. Also, the \glsxtrlong{SlReLU} (\glsxtrshort{SlReLU}; see \cite{Kunc2024Decades} \cref{sec:slrelu}) has a slope controlling parameter in a similar manner as the \gls{LReLU} (and its variants) but for positive inputs. Its \gls{TAAF} equivalent is $\alpha = a$, $\beta = 1$, $\gamma = 0$, $\delta = 0$ and $f(z) = \mathrm{ReLU}(z)$. 
There are several \glsxtrlongpl{AF} that use a parameter that modifies the range of the output of an \glsxtrlong{AF}. One of them is the E-swish \cite{Alcaide2018} (see \cite{Kunc2024Decades} \cref{sec:eswish}) which adds a parameter $a$ that is the equivalent of the \gls{TAAF}'s parameter $\alpha$ --- the E-swish is a special case of \gls{TAAF} if $\alpha=a$, $\beta = 1$, $\gamma = 0$, $\delta =0$ and $f(z) =z\cdot\sigma(z)$. The \gls{SGELU} (see \cite{Kunc2024Decades} \cref{sec:sgelu}) also uses a parameter $a$ for vertical scaling that controls the slope of the activation. While the parameter is fixed and nonadaptive, it can be tuned to reach better performance\cite{Yu2019Symmetrical}. The \gls{SGELU} can be considered as a special case of \gls{TAAF} with fixed parameters: $\alpha=a$, $\beta = 1$, $\gamma = 0$, $\delta = 0$, and $f(z) = z \cdot \mathrm{erf}\left(\frac{z}{\sqrt{2}}\right)$, where $\mathrm{erf}\left(x\right)$ is the Gauss error function.

The \gls{comb-H-sine} activation (see \cite{Kunc2024Decades} \cref{sec:combhsine}) uses a fixed parameter $a$ for input scaling; it can be considered as a special case of \glspl{TAAF} with $\alpha = 1$, $\beta = a$, $\gamma = 0$, $\delta=0$, and $f(z) = \sinh\left(z\right) + \sinh^{-1}\left(z\right)$.

The \gls{DRLU} adds a parameter for horizontal shifting but this time it is a fixed predefined parameter $a$; therefore, it can be considered to be nontrainable equivalent of \gls{TAAF} with $\alpha = 1$, $\beta=1$, $\gamma = a$, $\delta = 0$, and $f(z) = \mathrm{ReLU}(z)$.

The \gls{DReLU} (see \cite{Kunc2024Decades} \cref{sec:drelu}) has a parameter $a$ that shifts the basic \gls{ReLU} both horizontally and vertically; it is \gls{TAAF} equivalent for $\alpha = 1$, $\beta=1$, $\gamma = -a$, $\delta = a$, and $f(z) = \mathrm{ReLU}(z)$. The only difference is the calculation of the value of $a$ as the midpoint of the range of input values for each batch instead of optimizing it with the rest of a network's parameters. On the other hand, the \gls{DisReLU} (see \cite{Kunc2024Decades} \cref{sec:disrelu}) employs the identical concept with a fixed, predefined parameter $a$ instead of input dependent value --- the other difference is that the parameter is defined with a negative sign. The \gls{DisReLU} with parameter $a$ is a special case of \gls{TAAF} with  $\alpha = 1$, $\beta=1$, $\gamma = a$, $\delta = -a$, and $f(z) = \mathrm{ReLU}(z)$.

While the Flatted-T Swish (see \cite{Kunc2024Decades} \cref{sec:fts}) is a bit more complicated than a \gls{ReLU} with additional parameters, it can also be considered as a special case of a \gls{TAAF} but with more complicated function $f$ --- $\alpha = 1$, $\beta=1$, $\gamma = 0$, $\delta = T$, and $f(z) =  \mathrm{ReLU}(z) \cdot \sigma(z)$, where $T$ is the only parameter of the Flatted-T Swish.

The \gls{PSoftplus} \glsxtrlong{AF} (see \cite{Kunc2024Decades} \cref{sec:psoftplus}) has two fixed parameters $a$ and $b$ for scaling and translation; it can be considered as a special case of the \gls{TAAF} with $\alpha = a$, $\beta = 1$, $\gamma = 0$, $\delta = -ab$, and the function $f$ is the \gls{softplus} activation (see \cite{Kunc2024Decades} \cref{sec:softplus}) --- $f(z) = \ln\left(\exp\left(z\right) + 1 \right)$.

The functions listed above are equivalent to \glspl{TAAF} during the test phase or \glspl{TAAF} with frozen, nonadaptive parameters. More Interestingly, many functions can be considered as a special case of \glspl{TAAF}, including the property of adaptive parameters. One such function is the \gls{FReLU} (see \cite{Kunc2024Decades} \cref{sec:frelu}), which introduces parameters $a_i$ and $b_i$ for controlling the vertical and horizontal translation --- the \gls{TAAF} equivalent is with $\alpha=1$, $\beta=1$, $\gamma = a_i$, $\delta = b_i$, and $f(z) = \mathrm{ReLU}(z)$. The \gls{ShiLU} (see \cite{Kunc2024Decades} \cref{sec:shilu}) is adaptive variant of \gls{ReLU} that has adaptive vertical scaling using parameter $a_i$  and vertical translation using parameter $b_i$; the \gls{TAAF} equivalent is $\alpha = a_i$, $\beta=1$, $\gamma = 0$, $\delta = b_i$, and $f(z) = \mathrm{ReLU}(z)$. 

The \gls{ABReLU} \cite{Dubey2021} (see \cite{Kunc2024Decades} \cref{sec:abrelu}) has a parameter $a_i$ for horizontal shifting of the function; it has the same function as the $\gamma$ in \glspl{TAAF} but its value is not optimized using gradient descent as in \glspl{TAAF} but rather is calculated as the average of input activation map for each neuron $i$. The \gls{ABReLU} is \gls{TAAF} equivalent for $\alpha = 1$, $\beta=1$, $\gamma = -a_i$, $\delta = 0$, and $f(z) = \mathrm{ReLU}(z)$.
The \gls{PPReLU} (see \cite{Kunc2024Decades} \cref{sec:pprelu}) is an adaptive variant of the \gls{SlReLU}. Similarly, the \gls{pLogish} \cite{Zhu2021Logish} is a special case of nonadaptive \glspl{TAAF}; the equivalent parameterization is $\alpha = \frac{a}{b}$, $\beta = b$, $\gamma = 0$, $\delta = 0$ and $f(z) = z\cdot\ln\left(1+\sigma\left(z\right)\right)$.

The \gls{AOAF} (see \cite{Kunc2024Decades} \cref{sec:aoaf}) has three parameters, fixed $b$ and $c$ and adaptive parameter $a_i$ that is calculated as the mean value of the inputs of neuron $i$; these parameters are used for translation of the \glsxtrlong{AF}. The \gls{AOAF} can be considered as a special case of \gls{TAAF} but with a different scheme for updating the value of its parameters --- $\alpha = 1$, $\beta=1$, $\gamma = -ba_i$, $\delta = ca_i$, and $f(z) = \mathrm{ReLU}(z)$.

The \gls{LeLeLU} (see \cite{Kunc2024Decades} \cref{sec:lelelu}) is a \gls{LReLU} with an added trainable parameter for scaling, thus its \gls{TAAF} parameter is simply $\alpha = a_i$, $\beta=1$, $\gamma = 0$, $\delta = 1$, and $f(z) = \mathrm{LReLU}(z)$.

The \gls{RMAF} (see \cite{Kunc2024Decades} \cref{sec:rmaf}) is a bit more complicated \glsxtrlong{AF} that has one adaptive parameter $a_i$ for vertical scaling and two fixed parameters $b$ and $c$. Since the parameters $b$ and $c$ are fixed, the \gls{RMAF} can be formulated using the \gls{TAAF} framework --- $\alpha=a_i$, $\beta=1$, $\gamma=0$, $\delta=0$, and $f(z) = \left[b\frac{1}{\left(0.25\left(1+\exp(-z)\right)+0.75\right)^c}\right]\cdot z$.

The \gls{RSign} (see \cite{Kunc2024Decades} \cref{sec:rsign}) is a sign function with horizontal shift; its \gls{TAAF} formulation is therefore $\alpha=1$, $\beta=1$, $\gamma = -a_c$, $\delta = 0$, and $f(z) = \mathrm{sgn}(z)$.
 
The \gls{paired_relu} (see \cite{Kunc2024Decades} \cref{sec:paired_relu}) is a vector \glsxtrlong{AF} that outputs two values instead of one; however, the same result can be obtained using two \glspl{TAAF} that takes the same preactivation as the input and whose output values are then concatenated. The \gls{paired_relu} has four parameters $a_i$, $b_i$, $c_i$, and $d_i$, --- one pair for each output value. In each pair, there is one parameter for horizontal scaling and one for horizontal translation. The \gls{TAAF} based equivalent is $\alpha_1 = 1$, $\beta_1=a_i$, $\gamma_1 = -b_i$, $\delta_1 = 0$, and $f_1(z) = \mathrm{ReLU}(z)$ for the first \gls{TAAF} and $\alpha_2 = 1$, $\beta_2=c_i$, $\gamma_2 = -d_i$, $\delta_2 = 0$, and $f_2(z) = \mathrm{ReLU}(z)$ for the second \gls{TAAF}.

Similar approach to the \gls{paired_relu} is the \gls{MBA} (see \cite{Kunc2024Decades} \cref{sec:mba}) which can be seen as multiple \glspl{TAAF} applied to the same preactivation; in that case, each of $K$ \glspl{TAAF} would be defined as $\alpha=1$, $\beta=1$, $\gamma=b_{i,k},k=1,\ldots,K$, $\delta=0$ and $f(z)$ can be any \glsxtrlong{AF} --- authors used the \gls{ReLU} activation.

The \gls{SvELU} and \gls{ShELU} and its parametric variants (see \cite{Kunc2024Decades} \cref{sec:shifted_elus}) introduce an additional parameter to the \gls{ELU} activation controlling the translation. The \gls{ShELU} introduces horizontal translation controlled by a fixed hyperparameter $b$; it is a \gls{TAAF} equivalent with $\alpha=1$ (the \gls{ELU}, however, has its own parameter $a$ for vertical scaling of the function for negative inputs), $\beta=1$, $\gamma=b$, $\delta=0$, and $f(z) = \mathrm{ELU}(z)$. The \gls{SvELU} introduces vertical translation instead of horizontal, it is a \gls{TAAF} equivalent with $\alpha=1$, $\beta=1$, $\gamma=0$, $\delta=b$, and $f(z) = \mathrm{ELU}(z)$. The parametric variant \gls{PShELU} combines the \gls{ShELU} with the \gls{PELU} and, as such, introduces two additional parameters controlling the slope $a_i$ and $b_i$; these parameters, together with the \gls{ShELU}'s translation parameter $c_i$ are adaptive. The exact \gls{TAAF} equivalent is $\alpha=a_i$, $\beta=\frac{1}{b_i}$, $\gamma=\frac{c_i}{b_i}$, $\delta=0$, and $f(z) = \mathrm{ELU}(z)$. While \citeauthor{Grelsson2018} did not formulate the parameteric equivalent of the \gls{SvELU}; it was formulated as the \gls{PSvELU} in \cref{sec:shifted_elus} in \cref{eq:psvelu} --- the \gls{TAAF} equivalent parameterization is $\alpha=a_i$, $\beta=\frac{1}{b_i}$, $\gamma = 0$, $\delta = c_i$, and $f(z) = \mathrm{ELU}(z)$. Similar \glspl{AF} were proposed as variants of the \gls{hard_tanh} \gls{AF} - the \gls{SvHardTanh} introduces a fixed parameter for vertical shifts while the \gls{ShHardTanh} introduces a fixed parameter for horizontal shifts. Their \gls{TAAF} equivalents are $\alpha=1$, $\beta=1$, $\gamma=-a$, $\delta=0$, and $f(z) = \mathrm{HardTanh}(z)$ for \gls{ShHardTanh} and $\alpha=1$, $\beta=1$, $\gamma=0$, $\delta=a$, and $f(z) = \mathrm{HardTanh}(z)$ for \gls{SvHardTanh}.

\Citeauthor{Adem2022PFELU} proposed a novel variant of the \gls{FELU} by just adding a trainable parameter for vertical translation to the original \gls{AF}; this is exactly what the \gls{TAAF} does. Note that the original \gls{FELU} is also adaptive and has its own scaling parameter $a_i$ and its relation to the \glspl{TAAF} is discussed in \cref{sec:related_activations}.

There is also an adaptive variant of \gls{hard_tanh} (see \cite{Kunc2024Decades} \cref{sec:adaptive_hardtanh}) that can be considered as a special case of \glspl{TAAF} but with only parameter adaptive and the other is epoch dependent with a predefined schedule; the \gls{TAAF} equivalent is $\alpha=1$, $\beta=a_t$, $\gamma=-a_tb$, $\delta=0$, and $f(z) = \mathrm{HardTanh}(z)$ where $a_t$ is the fixed parameter scheduled for each epoch $t$ and $b$ is optimized along with other parameters as is usual for \glspl{TAAF}.

One of the \glsxtrlong{AAF} proposed earliest is the sigmoid function with shape autotuning (see \cite{Kunc2024Decades} \cref{sec:sigmoid_afs}, \cref{eq:yamada1992}). This function uses a single adaptive parameter $a \in (0, \infty)$ for controlling both the output range and the vertical scaling of the function; its equivalent within the \gls{TAAF} framework is $\alpha=a$, $\beta=-a$, $\gamma=0$, $\delta=0$, and $f(z) = 2\frac{1-\exp\left(-z\right)}{\left(1+\exp\left(-z\right)\right)}$. This approach was further extended into a \gls{generalized hyperbolic tangent} (see \cite{Kunc2024Decades} \cref{sec:generalized_hyperbolic_tangent}), which separates the parameters for controlling the amplitude and the vertical scaling into $a_i$ and $b_i$, which are adaptive parameters for each neuron $i$. The \gls{TAAF} equivalent is $\alpha=a_i$, $\beta=-b_i$, $\gamma=0$, $\delta=0$, and $f(z) = \frac{1-\exp\left(-z\right)}{\left(1+\exp\left(-z\right)\right)}$.

A predecessor of \glspl{TAAF} called trainable amplitude (see \cite{Kunc2024Decades} \cref{sec:sigmoid_afs}, \cref{eq:trainable_amplitude}) introduces two additional adaptive parameters to any inner \glsxtrlong{AF} $g(z)$; these two parameters $a_i$ and $b_i$ control vertical scaling and translation for each neuron $i$. Another general class of transformation of any \glsxtrlongpl{AF} was published in \cite{Jagtap2020} concurrently with our research \cite{Kunc2021_preprint} --- the class of slope varying \glsxtrlongpl{AF}. This class adds a single adaptive parameter $a$ to any \glsxtrlong{AF} $g(z)$ allowing for horizontal scaling of the function; it is equivalent to a \gls{TAAF} with $\alpha=1$, $\beta=a$, $\gamma=0$, $\delta=0$, and $f(z) = g(z)$. This general approach was preceded by a special cases called \gls{SVAF} (see \cite{Kunc2024Decades} \cref{sec:svaf}) that uses hyperbolic tangent function as the inner activation $f(z)$ and \gls{ASSF} (see \cite{Kunc2024Decades} \cref{sec:assf}) that uses \gls{logisticsigmoid} as the inner activation. The \gls{psigmoid} (see \cite{Kunc2024Decades} \cref{sec:psigmoid}) is another \gls{AAF} with scaling parameters. Unlike the \gls{SVAF}, the \gls{psigmoid} has both vertical and horizontal scaling parameters. Interestingly, only the vertical scaling parameter $a_i$ is local for each neuron or channel --- the horizontal scaling parameter $b$ is global. It can be considered as a special case of \glspl{TAAF} with some parameters shared and $\alpha=a_i$, $\beta=b$, $\gamma=0$, $\delta=0$, and $f(z_i) = \sigma(z_i)$. Another special case is the \gls{swish} (see \cite{Kunc2024Decades} \cref{sec:swish}), an adaptive variant of the later proposed SiLU activation. The \gls{swish} uses parameter $a_i$ for horizontal scaling; its \gls{TAAF} equivalent is $\alpha=1$, $\beta=a_i$, $\gamma=0$, $\delta=0$, and $f(z) = z \cdot \sigma(z)$. Another adaptive \gls{SiLU} variant is the \gls{AHAF} that employs both vertical and horizontal scaling; its \gls{TAAF} equivalent is, therefore, $\alpha=a_i$, $\beta=b_i$, $\gamma=0$, $\delta=0$, and $f(z) = z \cdot \sigma(z)$. The adaptive slope hyperbolic tangent (see \cite{Kunc2024Decades} \cref{sec:astanh}) is an adaptive function with horizontal scaling using parameter $a_i$ with the \gls{TAAF} parameterization $\alpha=1$, $\beta=_ai$, $\gamma=1$, $\delta=1$, $f(z) = \tanh(z)$. The \gls{PSTanh} (see \cite{Kunc2024Decades} \cref{sec:pstanh}) is an \glsxtrlong{AAF} that is a  cross between the adaptive slope hyperbolic tangent and the slope hyperbolic tangent. The \gls{PSTanh} has two scaling parameters --- $a_i$ for vertical scaling, $b_i$ for horizontal scaling; the \gls{TAAF} equivalent parameterization is $\alpha=a_i$, $\beta=b_i$, $\gamma = 0$, $\delta = 0$, and $f(z) =z \cdot \left(1+\tanh\left(z\right)\right)$. Similarly, the simpler \gls{SSinH} (see \cite{Kunc2024Decades} \cref{sec:ssinh}) has also two scaling parameters $a_i$ and $b_i$ and its equivalent \gls{TAAF} parameterization is $\alpha=a_i$, $\beta=b_i$, $\gamma = 0$, $\delta = 0$, and $f(z) =\sinh\left(z\right)$. Another scaled \gls{AF} is the \gls{SExp} which uses exponential instead of the $\sinh$ function; its \gls{TAAF} equivalent is $\alpha=a_i$, $\beta=b_i$, $\gamma = 0$, $\delta = 0$, and $f(z) =\exp\left(z\right)-1$.

Another sigmoid-based adaptive function that can be formulated within the \gls{TAAF} framework is the \gls{PFTS} (see \cite{Kunc2024Decades} \cref{sec:pfts}) that is the combination of a \gls{ReLU} and sigmoid activation with an adaptive parameter $T_i$ for vertical translation --- it is an adaptive variant of the \gls{FTS} (see \cite{Kunc2024Decades} \cref{sec:fts}). The \gls{TAAF} equivalent of \gls{PFTS} is $\alpha=1$, $\beta=1$, $\gamma=1$, $\delta=T_i$, and $f(z) = \mathrm{ReLU}(z) \cdot \sigma(z)$.

The \gls{parameterized_softplus} (see \cite{Kunc2024Decades} \cref{sec:parameterized_softplus}) has a parameter $a_i$ for controlling vertical shift of the \glsxtrlong{AF}; it is defined as $\alpha=1$, $\beta=1$, $\gamma=0$, $\delta=-a_i$, and $f(z) = \ln\left(1+\exp(z)\right)$ within the \gls{TAAF} framework albeit with the limiation of $\delta \in [-1, 0]$. 
The summary of \glsxtrlongpl{AF} found in the literature that can be formulated as special cases of \glspl{TAAF} is in \cref{tab:act_taaf_special_cases}.

The \gls{scaledlogisticsigmoid} (see \cite{Kunc2024Decades} \cref{sec:scaled_logistic_sigmoid})  is an adaptive function that is a special case of previously proposed \gls{NAF} (see \cite{Kunc2024Decades} \cref{sec:naf}) that has parameters $a_i$ and $b_i$ for controlling the vertical and horizontal scale of the function; its \gls{TAAF} equivalent is $\alpha=a_i$, $\beta=b_i$, $\gamma=0$, $\delta=0$, and $f(z) = \frac{1}{1+\exp\left(-z\right)}$.

A different approach where only the \glsxtrlongpl{AF} are trained, and the networks are kept randomly initialized is presented in \cite{Erturul2018} where five different \glsxtrlongpl{AAF} with two parameters $a_i$ and $b_i$ each were used. Four of these \glslink{AF}{activation} can be formulated within the \gls{TAAF} framework. The activation from \cref{eq:Erturul2018_af1} is equivalent to \gls{TAAF} with $\alpha=1$, $\beta=a_i$, $\gamma=b_i$, $\delta=0$, and $f(z) = \frac{1}{1+\exp\left(-z\right)}$.  The activation from \cref{eq:Erturul2018_af2} is equivalent to \gls{TAAF} with $\alpha=1$, $\beta=a_i$, $\gamma=b_i$, $\delta=0$, and $f(z) = \sin(z)$. And finally, the activation from \cref{eq:Erturul2018_af3} is equivalent to \gls{TAAF} with $\alpha=1$, $\beta=a_i$, $\gamma=-a_ib_i$, $\delta=0$, and $f(z) = \exp\left(-||z||\right)$. The activation from \cref{eq:Erturul2018_af4} could also be formulated within the \gls{TAAF} framework even though it is only a step function with a variable threshold that is determined by two parameters --- $\alpha=1$, $\beta=a_i$, $\gamma=b_i$, $\delta=0$, and 
\begin{equation}
    f(z) = \begin{cases}
        1, \quad & z \leq 0, \\
        0, \quad & \text{otherwise}.
     \end{cases} 
\end{equation}
 The final activation from \cite{Erturul2018} shown in \cref{eq:Erturul2018_af5} cannot be formulated within the \gls{TAAF} framework as only the parameter $a_i$ has an equivalent parameter within the \gls{TAAF} framework.

\begin{landscape}
    %V: see https://tex.stackexchange.com/a/599702 for the trick
    %\setlength\LTleft{-1cm}
    \begin{longtable}{ p{3cm}| c | c | c | c | p{1.5cm} | c | c | c | c | c | p{3.7cm} }
        activation & year & section in \cite{Kunc2024Decades}& source & adap. & param. & $\alpha$ & $\beta$ & $\gamma$ & $\delta$ & $f(z)$ & note \\
        \hline
        scaled hyperbolic tangent & \citeyear{Lecun1998} & \ref{sec:stanh} & \cite{Lecun1998} & \xmark & $a$, $b$ & $a$ & $b$ & $0$ & $0$ & $\tanh(z)$ & \\
        \gls{E-Tanh} & \citeyear{Kalaiselvi2022} & \ref{sec:etanh} & \cite{Kalaiselvi2022} & \xmark & $a$ & $a$ & $1$ & $0$ & $0$ & $\exp(z)\tanh(z)$ & \\
        \gls{SSS} & \citeyear{Arai2018} & \ref{sec:sss} & \cite{Arai2018} & \xmark & $a$, $b$ & $1$ & $a$ & $-ab$ & $0$ & $\sigma(z)$ & \\
        \gls{VSF} & \citeyear{Han1995} & \ref{sec:vsf} & \cite{Han1995} & \xmark & $a$, $b$, $c$ & $a$ & $b$ & $0$ & $-c$ & $\sigma(z)$ & \\
        \gls{SlReLU} & \citeyear{Seo2017} & \ref{sec:slrelu} & \cite{Seo2017} & \xmark & $a$ & $a$ & $1$ & $0$ & $0$ & $\mathrm{ReLU}(z)$ & \\
        \gls{pLogish} & \citeyear{Zhu2021Logish} & \ref{sec:plogish} & \cite{Zhu2021Logish} & \xmark & $a$, $b$ & $\frac{a}{b}$ & $b$ & $0$ & $0$ & $z\cdot\ln\left(1+\sigma\left(z\right)\right)$ & \\
        E-Swish & \citeyear{Alcaide2018} & \ref{sec:eswish} &\cite{Alcaide2018} & \xmark & $a$ & $a$ & $1$ & $0$ & $0$ & $z\cdot\sigma(z)$ & \\    
        \gls{ABReLU} & \citeyear{Dubey2021} & \ref{sec:abrelu} & \cite{Dubey2021} & \cmark &   $a_i$  &  $1$ & $1$ & $-a_i$ & $0$ & $\mathrm{ReLU}(z)$  & $a_i$ calculated as the average of a neuron's input map \\
        \gls{PPReLU} & \citeyear{Dai2022} & \ref{sec:pprelu} & \cite{Dai2022} & \cmark & $a$ & $a$ & $1$ & $0$ & $0$ & $\mathrm{ReLU}(z)$ & \\
        \gls{DRLU} & \citeyear{Shan2022} & \ref{sec:drlu} & \cite{Shan2022} & \xmark &   $a$  &  $1$ & $1$ & $a$ & $0$ & $\mathrm{ReLU}(z)$  & \\
        \gls{AOAF} & \citeyear{Jiang2022} & \ref{sec:aoaf} & \cite{Jiang2022} & \cmark &  $a_i$, $b$, $c$  & $1$ & $1$ & $-ba_i$ & $ca_i$ & $\mathrm{ReLU}(z)$   &  $a_i$ calculated as the average of a neuron's input map  \\
        \gls{DReLU} & \citeyear{Si2018} & \ref{sec:drelu} & \cite{Si2018} & \cmark & $a$  &  $1$ & $1$ & $-a$ & $a$ & $ \mathrm{ReLU}(z) $  &  $a$ calculated as the midpoint of range of input values for each batch \\
        \gls{DisReLU} & \citeyear{Macdo2019} & \ref{sec:disrelu} & \cite{Macdo2019} & \xmark &  $a$  &  $1$ & $1$ & $a$ & $-a$ & $\mathrm{ReLU}(z)$   &   \\
        Flatted-T Swish & \citeyear{Chieng2018} & \ref{sec:fts} & \cite{Chieng2018} & \xmark &  $T$  &   $1$ & $1$ & $0$ & $T$ & $ \mathrm{ReLU}(z) \cdot \sigma(z) $    \\
        \gls{PSoftplus} & \citeyear{Sun2019} & \ref{sec:psoftplus} & \cite{Sun2019} & \xmark &  $a$, $b$  & $a$ & $1$ & $0$ & $-ab$ & $\ln\left(\exp\left(z\right) + 1 \right)$  &   \\
        \gls{SGELU} & \citeyear{Yu2019Symmetrical} & \ref{sec:sgelu} & \cite{Yu2019Symmetrical} & \xmark & $a$  &  $a$ & $1$& $0$ & $0$ & $ z \cdot \mathrm{erf}\left(\frac{z}{\sqrt{2}}\right) $    \\
        \gls{comb-H-sine} & \citeyear{Vijayaprabakaran2022} & \ref{sec:combhsine} & \cite{Vijayaprabakaran2022} & \xmark &  $a$  & $1$ & $a$ & $0$ & $0$ & $\sinh\left(z\right) + \sinh^{-1}\left(z\right)$   &   \\
        \gls{FReLU}  & \citeyear{Qiu2018FReLU} & \ref{sec:frelu} & \cite{Qiu2018FReLU} & \cmark &  $a_i$, $b_i$  & $1$ & $1$ & $a_i$ & $b_i$ & $\mathrm{ReLU}(z)$   &   \\
        \gls{ShiLU}  & \citeyear{Pishchik2023} & \ref{sec:shilu} & \cite{Pishchik2023} & \cmark &  $a_i$, $b_i$  & $a_i$ & $1$ & $0$ & $b_i$ & $\mathrm{ReLU}(z)$   &   \\
        \gls{LeLeLU}  & \citeyear{Maniatopoulos2021} & \ref{sec:lelelu} & \cite{Maniatopoulos2021} & \cmark &  $a_i$  & $a_i$ & $1$ & $0$ & $0$ & $\mathrm{LReLU}(z)$   &   \\
        \gls{paired_relu} & \citeyear{Tang2018} & \ref{sec:paired_relu} & \cite{Tang2018} & \cmark &   $a_i$, $b_i$, $c_i$, $d_i$ & $1$ & $a_i$, $c_i$ & $b_i$, $d_i$ & $0$ & $\mathrm{ReLU}(z)$   & concatenation of two \glspl{TAAF}  \\
        \gls{RMAF}  & \citeyear{Yu2020} & \ref{sec:rmaf} & \cite{Yu2020} & \cmark &  $a_i$, $b$, $c$  & $a_i$ & $1$ & $0$ & $0$ & $\left[\frac{b\cdot z}{\left(0.25\left(1+\exp(-z)\right)+0.75\right)^c}\right] $   &  $b$ and $c$ are fixed \\
        \gls{RSign} & \citeyear{Liu2020ReActNet} & \ref{sec:rsign} & \cite{Liu2020ReActNet} & \cmark & $a_c$ & $1$ & $1$ & $-a_c$ & $0$ & $\mathrm{sgn}(z)$ & \\
        \gls{RPReLU} & \citeyear{Liu2020ReActNet} & \ref{sec:rprelu} & \cite{Liu2020ReActNet} & \cmark & $a_c$, $b_c$, $c_c$ & $1$ & $1$ & $-a_c$ & $b_c$ & $\mathrm{PReLU}(z)$ & $c_c$ is the parameter from  \gls{PReLU}\\
        \gls{ShELU} & \citeyear{Grelsson2018} & \ref{sec:shifted_elus} & \cite{Grelsson2018} & \xmark &  $a$, $b$  & $1$ & $1$ & $b$ & $0$ & $\mathrm{ELU}(z)$   & $a$ is fixed parameter of the inner function $f$  \\
        \gls{SvELU} & \citeyear{Grelsson2018} & \ref{sec:shifted_elus} & \cite{Grelsson2018} & \xmark &  $a$, $b$  & $1$ & $1$ & $0$ & $b$ & $\mathrm{ELU}(z)$   &   $a$ is fixed parameter of the inner function $f$  \\
        \gls{PShELU} & \citeyear{Grelsson2018} & \ref{sec:shifted_elus} & \cite{Grelsson2018} & \cmark &  $a_i$, $b_i$, $c_i$  & $a_i$ & $\frac{1}{b_i}$ & $\frac{c_i}{b_i}$ & $0$ & $\mathrm{ELU}(z)$   &   \\
        \gls{PSvELU} & --- & \ref{sec:shifted_elus} & --- & \cmark &  $a_i$, $b_i$, $c_i$  & $a_i$ & $\frac{1}{b_i}$ & $0$ & $c_i$ & $\mathrm{ELU}(z)$   &  proposed in \cref{sec:shifted_elus} \\
        \gls{ShHardTanh} & \citeyear{Kim2021} & \ref{sec:shifted_hardtanh} & \cite{Kim2021} & \xmark &  $a$ & $1$ & $1$ & $-a$ & $0$ & $\mathrm{HardTanh}(z)$   &  \\
        \gls{SvHardTanh} & \citeyear{Kim2021} & \ref{sec:shifted_hardtanh} & \cite{Kim2021} & \xmark &  $a$ & $1$ & $1$ & $0$ & $a$ & $\mathrm{HardTanh}(z)$   &  \\
        \gls{PFELU} & \citeyear{Adem2022PFELU} & \ref{sec:pfelu} & \cite{Adem2022PFELU} & \cmark &  $b$ & $1$ & $1$ & $0$ & $b$ & $\mathrm{FELU}(z)$   & The \gls{FELU} is adaptive and has its own parameter $a$ \\
        Adaptive \gls{hard_tanh}  & \citeyear{Liu2021Adaptive} & \ref{sec:adaptive_hardtanh} & \cite{Liu2021Adaptive} & \cmark &  $a_t$, $b$  & $1$ & $a_t$ & $-a_tb$ & $0$ & $\mathrm{HardTanh}(z)$   &   \\
        sigmoid with shape autotuning & \citeyear{Yamada1992} & \ref{sec:sigmoid_afs} & \cite{Yamada1992} & \cmark &  $a$  & $a$ & $-a$ & $0$ & $0$ & $2\frac{1-\exp\left(-z\right)}{\left(1+\exp\left(-z\right)\right)}$   &   \\
        \gls{generalized hyperbolic tangent} & \citeyear{Chen1996} & \ref{sec:generalized_hyperbolic_tangent} & \cite{Chen1996} & \cmark &  $a_i$, $b_i$  & $a_i$ & $-b_i$ & $0$ & $0$ & $\frac{1-\exp\left(-z\right)}{\left(1+\exp\left(-z\right)\right)}$   &   \\
        trainable amplitude & \citeyear{Trentin2001} & \ref{sec:trainable_amplitude} & \cite{Trentin2001} & \cmark &  $a_i$, $b_i$ & $a_i$ & $1$ & $0$ & $b_i$ & $g(z) $   & general approach allowing for any inner function  \\
        \gls{LAAF} & \citeyear{Jagtap2020} & \ref{sec:laaf} & \cite{Jagtap2020} & \cmark &  $a$  & $1$ & $a$ & $0$ & $0$ & $g(z) $   &  general approach allowing for any inner function \\
        \gls{SVAF} & \citeyear{Bai2009} & \ref{sec:svaf} & \cite{Bai2009} & \cmark &  $a$  & $1$ & $a$ & $0$ & $0$ & $\tanh(z)$   &   \\
        \gls{ASSF} & \citeyear{Nawi2009} & \ref{sec:assf} & \cite{Nawi2009} & \cmark &  $a$  & $1$ & $a$ & $0$ & $0$ & $\sigma(z)$   &   \\
        \gls{psigmoid} & \citeyear{Ying2021} & \ref{sec:psigmoid} & \cite{Ying2021} & \cmark &  $a_i$, $b$  & $a_i$ & $b$ & $0$ & $0$ & $\sigma(z)$   &  $b$ is a global parameter \\
        \gls{swish} & \citeyear{Ramachandran2017} & \ref{sec:swish} & \cite{Ramachandran2017} & \cmark &  $a_i$  & $1$ & $a_i$ & $0$ & $0$ & $ z \cdot \sigma(z)$   &   \\
        \gls{AHAF} & \citeyear{Bodyanskiy2022} & \ref{sec:ahaf} & \cite{Bodyanskiy2022} & \cmark &  $a_i$, $b_i$  & $a_i$ & $b_i$ & $0$ & $0$ & $ z \cdot \sigma(z)$   &   \\
        \gls{PFTS} & \citeyear{Chieng2020} & \ref{sec:pfts} & \cite{Chieng2020} & \cmark &  $T_i$  & $1$ & $1$ & $0$ & $T_i$ & $\mathrm{ReLU}(z) \cdot \sigma(z)$   &   \\
        Adaptive slope hyperbolic tangent & \citeyear{Kapoor2021} & \ref{sec:astanh} & \cite{Kapoor2021} & \cmark &  $a_i$  & $1$ & $a_i$ & $0$ & $0$ & $\tanh(z)$   &   \\
        \gls{PSTanh} & \citeyear{Adu2021} & \ref{sec:pstanh} & \cite{Adu2021} & \cmark &  $a_i$, $b_i$  & $a_i$ & $b_i$ & $0$ & $0$ & $z \cdot \left(1+\tanh\left(z\right)\right)$   &   \\
        \gls{SSinH} & \citeyear{Husain2021} & \ref{sec:ssinh} & \cite{Husain2021} & \cmark &  $a_i$, $b_i$  & $a_i$ & $b_i$ & $0$ & $0$ & $\sinh\left(z\right)$   &   \\
        \gls{SExp} & \citeyear{Husain2021} & \ref{sec:sexp} & \cite{Husain2021} & \cmark &  $a_i$, $b_i$  & $a_i$ & $b_i$ & $0$ & $0$ & $\exp\left(z\right)-1$   &   \\
        \gls{parameterized_softplus} & \citeyear{Vargas2023} & \ref{sec:parameterized_softplus} & \cite{Vargas2023} & \cmark &  $a_i$  & $1$ & $1$ & $0$ & $-a_i$ & $\ln\left(1+\exp(z)\right)$   &  $\delta \in [-1,0]$ \\
        \gls{scaledlogisticsigmoid} & \citeyear{Tezel2007} & \ref{sec:scaled_logistic_sigmoid} & \cite{Tezel2007} & \cmark &  $a_i$, $b_i$  & $a_i$ & $b_i$ & $0$ & $0$ & $\frac{1}{1+\exp\left(-z\right)}$   & special case of \gls{NAF}  \\
        \gls{MBA} & \citeyear{Li2016} & \ref{sec:mba} & \cite{Li2016} & \cmark &  $b_{i,k}, k=1,\ldots,K$  & $1$ & $1$ & $b_{i,k}$ & $0$ & $g(z)$   &  general approach allowing for any inner function; $K$ \glspl{TAAF} applied to same preactivation  \\
        \cref{eq:Erturul2018_af1} & \citeyear{Erturul2018} & \ref{sec:rand_nn_trainable_afs} & \cite{Erturul2018} & \cmark &  $a_i$, $b_i$  & $1$ & $-a_i$ & $b_i$ & $0$ & $\frac{1}{1+\exp\left(-z\right)}$   &  unnamed AF  \\
        \cref{eq:Erturul2018_af2} & \citeyear{Erturul2018} & \ref{sec:rand_nn_trainable_afs} & \cite{Erturul2018} & \cmark &  $a_i$, $b_i$  & $1$ & $a_i$ & $b_i$ & $0$ & $\sin(z)$   &  unnamed AF  \\
        \cref{eq:Erturul2018_af3} & \citeyear{Erturul2018} & \ref{sec:rand_nn_trainable_afs} & \cite{Erturul2018} & \cmark &  $a_i$, $b_i$  & $1$ & $a_i$ & $-a_ib_i$ & $0$ & $\exp\left(-||z||\right)$   &  unnamed AF  \\
        \cref{eq:Erturul2018_af4} & \citeyear{Erturul2018} & \ref{sec:rand_nn_trainable_afs} & \cite{Erturul2018} & \cmark &  $a_i$, $b_i$  & $1$ & $a_i$ & $b_i$ & $0$ & $\begin{cases}
            1, \quad & z \leq 0, \\
            0, \quad & \text{otherwise},
         \end{cases}$   &  unnamed AF  \\
         \hline
        \caption[Activation functions as special cases of TAAFs]{\textbf{Activation functions as special cases of \glspl{TAAF}} \\ \Glsxtrlongpl{AF}that can be formulated within the \gls{TAAF} framework as listed in \cite{Kunc2024Decades}. The columns $\alpha$, $\beta$, $\gamma$, $\delta$ and $f(z)$ show the equivalent formulation within the \gls{TAAF} framework.}
        \label{tab:act_taaf_special_cases}
    \end{longtable}
 \end{landscape}

\section{Activations related to TAAFs}
\label{sec:related_activations}
 Some of the \glsxtrlongpl{AF} proposed in literature employ similar concepts as the \glspl{TAAF} but cannot be considered to be a special case of the \glspl{TAAF}. Nevertheless, the motivation for the concepts remains similar as for \glspl{TAAF}. One such example is the \gls{improvedlogisticsigmoid} (see \cite{Kunc2024Decades} \cref{sec:improved_logistic_sigmoid}) that uses a fixed parameter $a$ for controlling the slope of the outermost pieces of the piecewise function. However, since the central part of the piecewise function is not subjected to the controllable slope, it cannot be formulated as a special case of a \gls{TAAF}. An \gls{AF} very similar to the \gls{improvedlogisticsigmoid} is the \gls{STAC-tanh} (see \cite{Kunc2024Decades} \cref{sec:stactanh}) --- the only difference is that it uses $\tanh$ instead of \gls{logisticsigmoid} and its parameters $a_i$ and $b_i$ are adaptive. While it has a different shape, the \gls{rsigelu} (see \cite{Kunc2024Decades} \cref{sec:rsigelu}) also has parameter $a$ for controlling the slope of the outermost components of the function. Similarly, the penalized hyperbolic tangent (see \cite{Kunc2024Decades} \cref{sec:penalized_hyperbolic_tangent}) has a fixed, slope-controlling parameter but only for negative inputs. The \gls{hexpo}  (see \cite{Kunc2024Decades} \cref{sec:hexpo}) is an \glsxtrlong{AF} with four fixed parameters that have similar functions as parameters $\alpha$ and $\beta$ in \glspl{TAAF}. The \gls{hexpo}  is a piecewise function that is defined separately for positive and negative inputs --- the parameter $a$ is the equivalent of \glspl{TAAF} $\alpha$ for positive inputs, and the parameter $c$ is the equivalent for negative inputs; similarly, parameters $b$ and $b$ are equivalents of $\frac{1}{\beta}$.

 Fixed slope controlling parameter in a piecewise function is also used in the \gls{LReLU}, \gls{VLReLU}, and \gls{OLReLU} (see \cite{Kunc2024Decades} \cref{sec:lrelu}) where the parameter controls the slope of the "leaky" part of the \glsxtrlong{AF} for negative inputs. Similarly, the \gls{SignReLU} (see \cite{Kunc2024Decades} \cref{sec:signrelu}) uses a parameter $a$ for controlling the slope for negative inputs; however, unlike the \gls{LReLU} and its variant, the function is not linear for negative inputs. The \gls{DLReLU} (see \cite{Kunc2024Decades} \cref{sec:dlrelu}) also has a fixed parameter for controlling the slope for negative inputs as \gls{LReLU} has, but it also has an additional parameter that scales the slope of the negative inputs further using the test error from the previous epoch. The \gls{RReLU} (see \cite{Kunc2024Decades} \cref{sec:rrelu}) uses a stochastic slope controlling parameter during training and a fixed for inference when it becomes the \gls{LReLU}. The \gls{EReLU} (see \cite{Kunc2024Decades} \cref{sec:erelu}) is similar to \gls{RReLU}, but it uses stochastic parameters for controlling the slope for positive inputs instead of negative inputs.  
 
 The \gls{ELU} (see \cite{Kunc2024Decades} \cref{sec:elu}) also has a parameter that linearly scales the function for negative inputs --- albeit since the function is controlled by an exponential, the main reason for the parameter is to control to which value the \gls{ELU} converges for inputs going to negative infinity. The \gls{SELU} (see \cite{Kunc2024Decades} \cref{sec:selu}) has two parameters $a$ and $b$ controlling the slope --- one ($a$) for the whole function and the other only for negative inputs ($b$). Since these parameters are fixed, it could be considered as a special case of \glspl{TAAF} with the first parameter equivalent to \gls{TAAF}'s $\alpha$ that is fixed and with the \gls{TAAF}'s inner activation $f(z)$ being parameterized with another parameter $b$. Similarly, its extension \gls{LSELU} (see \cite{Kunc2024Decades} \cref{sec:lselu}) has one parameter for controlling the slope for all inputs; however, the \gls{LSELU} is a sum of an \gls{ELU} and linear function for negative inputs and slope of each component is controlled separately by parameters $b$ and $c$. The \gls{sSELU} (see \cite{Kunc2024Decades} \cref{sec:sselu}) has two parameters $a$ and $b$ for vertically scaling the function separately for negative and positive inputs; it also has a parameter $c$ for horizontally scaling the function for negative inputs similarly as does $\beta$ in \glspl{TAAF}.
 The \gls{rsigelud} (see \cite{Kunc2024Decades} \cref{sec:rsigelud}) also has two parameters for controlling the slope of individual components of the function; it has parameter $a$ for controlling the slope of the exponential component for inputs above one, and parameter $b$ for controlling the slope for negative inputs. However, since it has no parameter for controlling the slope for inputs in the interval $[0,1]$, where it is defined as a linear function, and since the parameter $a$ does not control the slope of the whole function for inputs above one but rather only weights one component of the function, it cannot be considered as a special case of a \gls{TAAF} but with different parameterization for positive and negative inputs as many other functions can.

 The \gls{softmodulust} (see \cite{Kunc2024Decades} \cref{sec:softmodulust}) uses a fixed, predefined parameter $a$ for scaling the input of the function similarly as the \gls{TAAF}'s parameter $\beta$ albeit in an inverse form --- $\beta \sim \frac{1}{a}$ --- and only for the input going to the hyperbolic tangent function. 

 The \gls{NReLU} (see \cite{Kunc2024Decades} \cref{sec:nrelu}) introduces a stochastic variant of the shift parameter $\gamma$ --- the mean value of the parameter is 0, and, therefore, it only introduces additive noise during training. The motivation behind \gls{NReLU} is different from the motivation of the \glspl{TAAF}, but nevertheless, the concept of the additive parameter to specific inputs resembles the \gls{TAAF}'s parameter $\gamma$. The \gls{RT-ReLU} (see \cite{Kunc2024Decades} \cref{sec:rtrelu}) also introduces stochastic translational parameter as the \gls{NReLU} but samples the parameters from different distributions. The \gls{ReSP} (see \cite{Kunc2024Decades} \cref{sec:resp}) also has a fixed parameter controlling the slope of the function for positive inputs only. On the other hand, the \gls{BLReLU} (see \cite{Kunc2024Decades} \cref{sec:blrelu}) has a fixed parameter controlling the slope for the negative inputs and also for inputs above a threshold predefined by another parameter similar to the \gls{improvedlogisticsigmoid}. 

 The Soft++ \glsxtrlong{AF} (see \cite{Kunc2024Decades} \cref{sec:softpp}) is a composition of a horizontally scaled \gls{softplus} activation using parameter $a$ and vertically scaled linear function using parameter $b$ with an additional fixed offset. While it cannot be considered as a special case of the \gls{TAAF} due to the composition of the two functions, the parameter $a$ has an identical role as the parameter $\beta$ in \glspl{TAAF}, the parameter $b$ scales the linear component similarly as parameter $\frac{1}{\alpha}$ and the linear offset can be defined as $\delta = -\ln(2)$.

 While the \glsxtrlong{AF} above do not have the parameters trainable --- some of them use different adaptive schemes --- there are also other \glsxtrlongpl{AF} that uses adaptive, trainable parameters similarly to \glspl{TAAF}. One of them is the \gls{PReLU} (see \cite{Kunc2024Decades} \cref{sec:prelu}) that is basically a \gls{LReLU}, but the parameter $a$ is adaptive. The \gls{TAAF}'s parameter $\alpha$ is the equivalent of \gls{PReLU}'s parameter $\frac{1}{a}$ but only for negative inputs; there is no adaptive scaling for positive inputs. The \gls{RT-PReLU} (see \cite{Kunc2024Decades} \cref{sec:rtprelu}) is the \gls{PReLU} but with additional stochastic parameter $b$ that is randomly sampled and that controls the threshold of the piecewise function.

 The \gls{PREU} (see \cite{Kunc2024Decades} \cref{sec:preu}) has vertical scaling for the whole function, and, therefore, its parameter $a$ is the direct equivalent of \gls{TAAF}'s parameter $\alpha$. However, it also introduces an equivalent for \gls{TAAF}'s parameter $\beta$ but only for negative inputs; therefore, it cannot be considered a special case of the \gls{TAAF}.

 The \gls{AReLU} (see \cite{Kunc2024Decades} \cref{sec:arelu}) has two adaptive scaling parameters $a_l$ and $b_l$  as it has separate scaling of positive and negative inputs. However, its difference from the \gls{TAAF} is much larger --- the parameter $b_l$ scaling positive inputs is transformed using the \gls{logisticsigmoid} into interval $[1,2]$ by $ \left(1+\sigma(b_l)\right)$ and the parameter $a_l$ for scaling negative inputs is clipped into interval $[0.01, 0.99]$.
 
The \gls{tanhLU} (see \cite{Kunc2024Decades} \cref{sec:tanhlu}) uses both vertical and horizontal scaling; however, since it has two components, it uses a separate parameterization for each of the components. The $\tanh$ component has a parameter $a_i$ for vertical scaling and a parameter $b_i$ for horizontal scaling, whereas the linear function has only a single parameter $c_i$ for scaling as there is no difference between vertical and horizontal scaling of linear functions.

 Separate adaptive parameters for controlling the slope are used in several \glsxtrlongpl{AAF}. One of the simplest examples is the \gls{DPReLU} (see \cite{Kunc2024Decades} \cref{sec:dprelu}), which is a piecewise linear function with one parameter controlling the slope for positive inputs and the other for negative inputs. The \gls{DPReLU} is extended by an adaptive parameter $m_i$ controlling vertical translation into the \gls{dual_line} \glsxtrlong{AF} (see \cite{Kunc2024Decades} \cref{sec:dual_line}). Since the translation parameter $m_i$ is shared by both piecewise components of the function, it is a direct equivalent of \gls{TAAF}'s parameter $\delta$.

 The \gls{PiLU} (see \cite{Kunc2024Decades} \cref{sec:pilu}) is another \gls{DPReLU} extension; it generalizes the \gls{dual_line} by adding any horizontal shift --- it has two parameters for vertical scaling (one for inputs below the threshold and one for inputs above the threshold) with similar function as the \gls{TAAF}'s $\alpha$ and one single parameter for the threshold which allows for the horizontal shift similarly as does the $\gamma$ in \glspl{TAAF}.

 Similarly as \glspl{TAAF} accept any inner \glsxtrlong{AF}, the DPAFs (see \cite{Kunc2024Decades} \cref{sec:dpaf}) extends the \gls{dual_line} concept to use any suitable inner \glsxtrlong{AF}; the \gls{DPAF} uses an inner \glsxtrlong{AF} $g(z_i)$ instead of the linear function from the \gls{dual_line}. It closely resembles the \gls{TAAF} with $\alpha$ applied only for positive inputs, $\beta=1$, $\gamma=0$ and $\delta=m_i$. The \gls{FPAF} (see \cite{Kunc2024Decades} \cref{sec:fpaf}) is very similar to \gls{DPAF}, but it allows for two different inner functions, one for positive inputs and the other for negative inputs. Each of the inner functions has its own adaptive parameter for vertical scaling, but unlike \gls{DPAF}, there is no adaptive translation parameter.

The \gls{EPReLU} (see \cite{Kunc2024Decades} \cref{sec:eprelu}) also has two separate parameters for positive and negative inputs that control the vertical scaling; however, only the parameter $a_i$ scaling the negative inputs is trainable; the parameter scaling the function for positive inputs is stochastic and sampled from a uniform distribution centered around 1 in each training epoch.

The \gls{PTELU} (see \cite{Kunc2024Decades} \cref{sec:ptelu}) behaves as linear function for positive inputs and as a special case of \gls{TAAF} for negative inputs with $\alpha = a_i$, $\beta=b_i$, $\gamma=0$, $\delta=0$, and $f(z) = \tanh(z)$ where $a_i$ and $b_i$ are trainable parameters for each neuron $i$. The later proposed \gls{TReLU} (see \cite{Kunc2024Decades} \cref{sec:trelu}) is identical to \gls{PTELU} but with fixed $\alpha = a_i = 1$ as it only has horizontal scaling for negative inputs.

The \gls{BLU} (see \cite{Kunc2024Decades} \cref{sec:blu}) is an \glsxtrlong{AF} that has two components --- nonlinear function with adaptive scaling parameter $a_i$ and a linear component. While the scaling parameter has a similar role as the \gls{TAAF}'s parameter $\alpha$, its values are limited to the range $[-1, 1]$.

The \gls{PELU} (see \cite{Kunc2024Decades} \cref{sec:pelu}) extends the \gls{ELU} by two parameters $a_i$ and $\frac{a_i}{b_i}$ controlling the slope --- separate parameters for positive and negative inputs --- but it also has a horizontal scaling parameter $\frac{1}{b_i}$ for the exponential part of the \gls{PELU} for negative inputs. The parameters $a_i$ and $b_i$ are formulated such that there is no non-differentiability at input $z=0$. Another \gls{ELU} extension \gls{FELU} (see \cite{Kunc2024Decades} \cref{sec:felu}) uses adaptive scaling parameter $a_i$ to control the soft saturation region for negative inputs. The \gls{MPELU} (see \cite{Kunc2024Decades} \cref{sec:mpelu}) outputs identity for positive inputs, but it outputs non-linearly transformed input for negative values that can be formulated within the \gls{TAAF} framework --- with $\alpha=a_i$, $\beta=b_i$, $\gamma=0$, $\delta=0$, and $f(z) = \exp(z)-1$ where $a_i$ and $b_i$ are \gls{MPELU}'s scaling parameters. The \gls{CELU} (see \cite{Kunc2024Decades} \cref{sec:celu}) is similar to \gls{MPELU} but it is reparameterized using a single parameter $a_i$ such that its derivative at $z=0$ is 1 --- the only difference from the \gls{MPELU} is that its \gls{TAAF} reformulation for negative part is $\alpha=a_i$ and $\beta = \frac{1}{a_i}$.

The \gls{PSELU} (see \cite{Kunc2024Decades} \cref{sec:pselu}), which is the adaptive variant of \gls{SELU}, has two trainable scaling parameters $a_i$ and $b_i$ that allow for vertical scaling of the function; the parameter $a_i$ is scaling the whole function and as such is the exact equivalent of the parameter $\alpha$ while the parameter $b_i$ scales the function only for negative inputs. The \gls{LPSELU} (see \cite{Kunc2024Decades} \cref{sec:lpselu})  is identical to \gls{PSELU}, but it adds a linear function for negative inputs to avoid small gradients --- this linear function has slope controlled by another parameter $c_i$; i.e., the function has two parameters that control the slope only for negative inputs, $b_i$ for the exponential part and $c_i$ for the linear part, the parameter $a_i$ controls the slope of the whole function. The \gls{LPSELURP} (see \cite{Kunc2024Decades} \cref{sec:lpselurp}) extends the \gls{LPSELU} by the additional parameter $m_i$ that controls the vertical translation of the whole function; this trainable parameter represents an exact equivalent of the \gls{TAAF}'s parameter $\delta$.

The \gls{PDELU} (see \cite{Kunc2024Decades} \cref{sec:pdelu}) introduces two parameters $a_i$ and $b$; while $a_i$ is an adaptive parameter that controls the scaling of the function for negative inputs, parameter $b$ is a fixed hyperparameter controlling the shape of the nonlinear part of the activation. Similarly, the \gls{T-swish} (see \cite{Kunc2024Decades} \cref{sec:tswish}) has parameters $a_i$ and $b_i$ for vertical and horizontal scaling only for negative inputs. 

The \gls{EELU} (see \cite{Kunc2024Decades} \cref{sec:eelu}) is an \glsxtrlong{AF} with a stochastic component for positive inputs and function scaling for negative inputs. The function is scaled using parameter $k_i$ for positive inputs, which is stochastic and is sampled from a Gaussian distribution with random variance (sampled from a uniform distribution) and is clipped into interval $[0,2]$. Adaptive parameters $a^c$ and $b^c$ are used for vertical and horizontal scaling of the function for negative inputs and are shared by all neurons in channel $c$; the function for negative inputs can be formulated within \gls{TAAF} framework using $\alpha=a^c$, $\beta=b^c$, $\gamma=0$, $\delta=0$, and $f(z) = \exp(z)-1$.

The \gls{scaled_softsign} (see \cite{Kunc2024Decades} \cref{sec:scaled_softsign}) is controlled by two adaptive parameters $a_i$ and $b_i$; however, only the $a_i$ has an equivalent within the \gls{TAAF} framework --- the parameter $a_i$ controls the vertical scale of the function and thus it is the equivalent of the parameter $\alpha$ of the \gls{TAAF} framework. The parameter $b_i$ controls the rate of transition between signs, and as such, it does not have an equivalent within the \gls{TAAF} framework.

The \gls{NAF} (see \cite{Kunc2024Decades} \cref{sec:naf}) consists of parts; each one has one parameter for controlling its vertical scale. The parts also have two additional parameters for controlling the horizontal scale similarly as does $\beta$ in the \glspl{TAAF} --- the first part has parameter $b ~ \beta^2$ while the second uses parameter $d$ that has the same function as $\beta$ without any non-linear transformation. Function similar to \gls{NAF} is the combination of \glsxtrlong{SLS-SS} (\glsxtrshort{SLS-SS}; see \cite{Kunc2024Decades} \cref{sec:scaled_logistic_sigmoid}) uses four parameters, one pair for controlling the horizontal and vertical scale of the \gls{logisticsigmoid} and the other pair controlling the horizontal and vertical scale of the sine function; the function can be seen also as the combination of two \gls{TAAF} based functions --- the first with $\alpha=a_i$, $\beta=b_i$, $\gamma=0$, $\delta=0$, $f(z) = \sin(z)$ and the second with $\alpha=c_i$, $\beta=d_i$, $\gamma=0$, $\delta = 0$, and $f(z) = \frac{1}{1+\exp\left(-z\right)}$.

The \gls{APLU} (see \cite{Kunc2024Decades} \cref{sec:aplu}) can be seen as sum of $S+1$ \gls{TAAF} based functions where the first function is just plain $\mathrm{ReLU}(z)$ while the others can be defined as \gls{TAAF} equvialents with $\alpha=a_i^s$, $\beta=1$, $\gamma=-b_i^s$, $\delta = 0$, and $f_s(z) = \mathrm{ReLU}(-z)$.

The \gls{MeLU} (see \cite{Kunc2024Decades} \cref{sec:melu}) is an approach with the same representation power as the \gls{APLU} but with a lower number of parameters; it consists of a sum of functions, each having its own trainable parameter for vertical scaling $a_{i,j}$.

Function combining sigmoid-like and \gls{ReLU} functions is the \gls{SReLU} (see \cite{Kunc2024Decades} \cref{sec:srelu}); it is a piecewise function with linear function in the middle and with two trainable determining thresholds limiting the middle identity segment; it also has two trainable parameters controlling the slope of the outermost segments. Similarly, the \gls{LinQ} (see \cite{Kunc2024Decades} \cref{sec:linq}) has one non-adaptive parameter for scaling the slope of the function but only for the parts that are outside the interval $[-2,2]$. The \gls{PLU} (see \cite{Kunc2024Decades} \cref{sec:plu}) can be considered as a special case of the \gls{SReLU} enforcing invertibility of the function; it has only one trainable parameter $a_i$ that determines the slope of two linear segments similarly as $\alpha$ does in \glspl{TAAF}. 
The \gls{AdaLU} (see \cite{Kunc2024Decades} \cref{sec:adalu}) is a piecewise linear function with adaptive parameters for controlling the slope and shifts of individual components.

The \gls{MTLU} (see \cite{Kunc2024Decades} \cref{sec:mtlu}) extends the \gls{SReLU} approach into more than three segments; each of the $K$ segments has a parameter $a_{i,k}, k=0,\ldots, K$ that controls the slope of the respective segment (a local equivalent of $\alpha$)and parameter $b_{i,k}, k=0,\ldots,K$ that controls its translation (a local equivalent of $\delta$); the segments are determined by parameters $c_{i,0}, \ldots, c_{i, K-1}$. The \gls{LuTU} (see \cite{Kunc2024Decades} \cref{sec:lutu}) is also a piecewise linear \glsxtrlong{AF} where each segment has adaptive slope and bias --- however, the function is defined by several anchor points instead of using direct equivalents of $\alpha$ and $\delta$ for each segment.

The \gls{maxout} (see \cite{Kunc2024Decades} \cref{sec:maxout}) returns a maximum of multiple linear functions; it can also be seen as returning maximum of $K$ \glspl{TAAF}; each with $\alpha=w_i^k$, $\beta=1$, $\gamma=0$, $\delta=b_i^k$, and $f(z)=z$, $k=1, \ldots, K$. 

The \gls{DY--ReLU} (see \cite{Kunc2024Decades} \cref{sec:dyrelu}) is a different approach compared to most of the adaptive functions in this list --- it uses a hyperfunction for computing the parameters of the \glsxtrlong{AF}. The \glsxtrlong{AF} itself is a piecewise linear function that is defined as the maximum of multiple linear functions --- it is a maximum of multiple independent \glspl{TAAF}, each with two parameters that are equivalent to the \gls{TAAF} parameters $\alpha$ and $\delta$.

A similar approach to the \gls{maxout} is the \gls{ABU} and its variants (see \cite{Kunc2024Decades} \cref{sec:abu}) --- using a weighted sum of \glsxtrlongpl{AF} instead of the maximum. This can be seen as a sum of \gls{TAAF} based functions when the weight $a_j,l$ is equivalent to the scaling parameter $\alpha$ for the relevant \gls{TAAF} with any inner activation $g_j(z)$. The formulation of \gls{ABU} as a sum of \gls{TAAF} is beneficial for the extended variant with additional bias parameter (see \cite{Kunc2024Decades} \cref{eq:abu_bias}) --- this \gls{ABU} is a sum of $n$ \gls{TAAF} based functions with $\alpha_{j}=a_{i,j}$, $\beta_j=1$, $\gamma_j = -b_{i,j}$, $delta_j = 0$ for any inner \glsxtrlong{AF}. There are other \gls{ABU} variants whose weights of individual inner \glsxtrlongpl{AF} have to sum up to 1 \cite{Manessi2018, Klabjan2019} or that are employing min--max scaling \cite{Klabjan2019}. Another \gls{ABU} variant called \gls{APAF} divides the output by the sum of the weighting coefficients --- the output is the weighted average of the inner \glsxtrlongpl{AF}. The \gls{GABU} (see \cite{Kunc2024Decades} \cref{sec:gabu}) is an \gls{ABU} variant that uses gating functions for obtaining the scaling parameters of individual inner \glsxtrlongpl{AF}. The \gls{SLAF} (see \cite{Kunc2024Decades} \cref{sec:slaf}) is a special case of \gls{ABU} that utilizes the increasing powers of the input as the individual inner \glsxtrlongpl{AF}. Similarly, the \gls{ChPAF} (see \cite{Kunc2024Decades} \cref{sec:chpaf}) and \gls{LPAF} (see \cite{Kunc2024Decades} \cref{sec:lpaf}) can be considered as \gls{ABU}, but the inner functions are Chebyshev and Legendre polynomials instead.

The \gls{SinLU} (see \cite{Kunc2024Decades} \cref{sec:sinlu}) uses vertical and horizontal scaling parameters $a_i$ and $b_i$ only for a single term in its definition that adds a sine function to the base linear function.

The \gls{KAF} (see \cite{Kunc2024Decades} \cref{sec:kaf}) is an \glsxtrlong{AF} that uses kernel expansion with a dictionary; however, since \citeauthor{Scardapane2019} used $D$ fixed dictionary points, it can also be viewed as a sum of individually scaled functions with parameters $a_{i,j}$,  $j=1,\ldots, D$.

The \gls{PAU} (see \cite{Kunc2024Decades} \cref{sec:pau}) extends the \gls{ABU} concept even further; a \gls{PAU} is basically a division of two SLAFs --- i.e., the \gls{PAU} is the division of two sums of individually transformed functions that are polynomials of increasing power. The \gls{ERA} (see \cite{Kunc2024Decades} \cref{sec:era}) is a function that is very similar to the \gls{PAU}; however, the \gls{ERA} is parameterized in such way that it can be rewritten using partial fractions reducing the number of operations --- this formulation however holds even less similarities with the \gls{TAAF} parameterization.

The \gls{MoGU} (see \cite{Kunc2024Decades} \cref{sec:mogu}) is, similarly to the \gls{ABU}, also a sum of individually transformed functions. However, unlike \gls{ABU}, the \gls{MoGU} uses more \gls{TAAF} parameters than just the $\alpha$. It can be defined as a sum of $n$ \gls{TAAF} based functions with $\alpha_j = \frac{a_{i,j}}{\sigma_{i,j}}$, $\beta_j = \frac{1}{\sigma_{i,j}}$, $\gamma_j = \frac{-\mu_{i,j}}{\sigma_{i,j}}$, $\delta=0$, and $f_j(z) = \frac{1}{\sqrt{2\pi}} \exp\left(-\frac{1}{2}\left(z\right)^2\right)$, $j=1,\ldots,n$. Similarly, the \gls{TCA} and \gls{TCAv2} (see \cite{Kunc2024Decades} \cref{sec:tca}) can be seen as a sum (\gls{TCA}) or a weighted average (\gls{TCAv2}) of $k$ \gls{TAAF} based functions. The \gls{TAAF} based functions are using parameters 
$\beta{i,j} = \exp\left(a_{i,j}\right)$ and $\gamma{i,j} = \exp\left(b_{i,j}\right)$ in \gls{TCA} and $\alpha{i,j} = \exp\left(a_{i,j}\right)$, $\beta{i,j} = \exp\left(b_{i,j}\right)$, and $\gamma{i,j} = \exp\left(c_{i,j}\right)$in \gls{TCAv2}. Note that the sum of the functions in \gls{TCAv2} is divided by $\sum_{j=1}^k \exp\left(a_{i,j}\right)$ to obtain the weighted average of the functions.

The \gls{MSAF} (see \cite{Kunc2024Decades} \cref{sec:msaf} is a sum of individually translated \glspl{logisticsigmoid}); it has a parameter for vertical translation $a$ and each \gls{logisticsigmoid} has another translation parameter $b_k$ for horizontal translation. These translations, however, seem to be predefined and nonadaptive.

Similarly, the \gls{FSA}  (see \cite{Kunc2024Decades} \cref{sec:fsa}) is also a sum of individually transformed functions; however, there are two different functions this time, and they are transformed using equivalents of both $\alpha$ and $\beta$ --- i.e., they have parameters for both horizontal and vertical scaling. Furthermore, there is also a single parameter $a_i$ that controls the vertical translation similarly to the parameter $\delta$ in \glspl{TAAF}.

The \gls{VAF} (see \cite{Kunc2024Decades} \cref{sec:vaf}) approach, published parallelly with the \glspl{TAAF}, uses a specially defined subnetwork instead of a simple \glsxtrlong{AF}; the resulting \glsxtrlong{AF} from the subnetwork is equivalent to the sum of \glspl{TAAF} in the most general sense --- it has all four \gls{TAAF} parameters in equivalent formulation and also allows for usage of any inner function. While the \gls{VAF} is more general than \gls{TAAF}, it also has significantly more parameters proportional to the size of the subnetwork.

The \cref{tab:act_related_activations} summarizes the \glsxtrlongpl{AF} that uses concepts that are related to those used in \glspl{TAAF}.

 \begin{landscape}
    \begin{longtable}{ p{3cm}| c | c | c | c |  p{2cm} | c | p{5cm}}
        activation & year & details in \cite{Kunc2024Decades} & source& adapt.& parameters & \gls{TAAF} equiv. & note \\
        \hline
        \gls{improvedlogisticsigmoid} & \citeyear{Qin2019} & \cref{sec:improved_logistic_sigmoid} & \cite{Qin2019} & \xmark & $a$, $b$ & $\alpha$ & controllable slope only for certain inputs \\
        \gls{STAC-tanh} & \citeyear{Zhang2021ANovel} & \cref{sec:stactanh} & \cite{Zhang2021ANovel} & \cmark & $a_i$, $b_i$ & $\alpha$ & controllable slope only for certain inputs determined by adaptive thresholds \\
        \gls{rsigelu} & \citeyear{Kiliarslan2021} & \cref{sec:rsigelu} & \cite{Kiliarslan2021} & \xmark &  $a$ & $\alpha$ & controllable slope only for certain inputs  \\
        penalized hyperbolic tangent & \citeyear{Xu2016} & \cref{sec:penalized_hyperbolic_tangent} &\cite{Xu2016} & \xmark & $a$ & $\alpha$ & controllable slope only for certain inputs \\
        \gls{hexpo}  & \citeyear{Kong2017} & \cref{sec:hexpo} & \cite{Kong2017} & \xmark & $a$, $c$; $b$, $d$ & $\alpha$; $\beta$ & different parameters for negative and positive inputs\\
        \gls{LReLU}  & \citeyear{Maas2013} & \cref{sec:lrelu} & \cite{Maas2013} & \xmark &  $a$  &  $\alpha$  &  controllable slope only for certain inputs \\
        \gls{VLReLU}  & \citeyear{Graham2014} & \cref{sec:lrelu} & \cite{Graham2014} & \xmark &   $a$  &  $\alpha$   & controllable slope only for certain inputs  \\
        \gls{OLReLU}  & \citeyear{Nayef2021} & \cref{sec:lrelu} & \cite{Nayef2021} & \xmark &    $a$  &  $\alpha$  & controllable slope only for certain inputs  \\
        \gls{DLReLU} & \citeyear{Hu2019} & \cref{sec:dlrelu} & \cite{Hu2019} & \xmark &  $ab_t$  &  $\alpha$  &  controllable slope only for certain inputs; slope controlled by a fixed parameter ($a$) but also using the test error from previous epoch ($b_t$) \\
        \gls{softmodulust} & \citeyear{VallsPrez2023} & \cref{sec:softmodulust} & \cite{VallsPrez2023} & \xmark &  $a$  &  $\beta$ & horizontal scaling only of the tanh component  \\
        \gls{SignReLU} & \citeyear{Lin2018} & \cref{sec:signrelu} & \cite{Lin2018} & \xmark &   $a$  &  $\alpha$  & controllable slope only for certain inputs  \\
        \gls{RReLU}  & \citeyear{Xu2015} & \cref{sec:rrelu} & \cite{Xu2015} & \xmark &    $a$  &  $\alpha$  & controllable slope only for certain inputs, stochastic  \\
        \gls{EReLU} & \citeyear{Jiang2018} & \cref{sec:erelu} & \cite{Jiang2018} & \xmark & $a$  &  $\alpha$ &  controllable slope only for certain inputs  \\
        \gls{NReLU} & \citeyear{Nair2010} & \cref{sec:nrelu} & \cite{Nair2010} & \xmark &  $a$  &  $\gamma$  & stochastic parameter with zero mean  \\
        \gls{RT-ReLU} & \citeyear{Cao2018Randomly} & \cref{sec:rtrelu} & \cite{Cao2018Randomly} & \xmark &  $a$  &  $\gamma$  & stochastic parameter with zero mean \\
        \gls{ReSP} & \citeyear{Xu2018Novel} & \cref{sec:resp} & \cite{Xu2018Novel} & \xmark & $a$  &  $\alpha$ &  controllable slope only for certain inputs \\
        \gls{BLReLU} & \citeyear{Liew2016} & \cref{sec:blrelu} & \cite{Liew2016} & \xmark &   $a$  &  $\alpha$ &  controllable slope only for certain inputs  \\
        \gls{ELU} & \citeyear{Clevert2015} & \cref{sec:elu} & \cite{Clevert2015} & \xmark &    $a$  &  $\alpha$  & controllable slope only for certain inputs  \\
        \gls{SELU} & \citeyear{Klambauer2017} & \cref{sec:selu} & \cite{Klambauer2017} & \xmark &    $a$, $b$  &  $\alpha$  & separately controllable slope for positive and negative inputs   \\
        \gls{LSELU} & \citeyear{Chen2021Redefining} & \cref{sec:lselu} & \cite{Chen2021Redefining} & \xmark &   $a$, $b$, $c$  &  $\alpha$ &  individual components has separate parameters for controlling the slope \\
        \gls{sSELU} & \citeyear{Chen2021Redefining} & \cref{sec:sselu} & \cite{Chen2021Redefining} & \xmark &  $a$, $b$, $c$  &  $\alpha$, $\beta$ &  individual components have separate parameters for controlling the slope \\
        \gls{rsigelud} & \citeyear{Kiliarslan2021} & \cref{sec:rsigelud} & \cite{Kiliarslan2021} & \xmark &  $a$, $b$  &  $\alpha$ &  individual components have separate parameters for controlling the slope  \\
        Soft++ & \citeyear{Ciuparu2020} & \cref{sec:softpp} & \cite{Ciuparu2020} & \xmark &  $a$, $b$  & $\alpha$, $\beta$  & one component is vertically scaled, the other horizontally scaled  \\
        \gls{PReLU} & \citeyear{He2015} & \cref{sec:prelu} & \cite{He2015} & \cmark &  $a$  &  $\alpha$ &  controllable slope only for certain inputs  \\ 
        \gls{RT-PReLU} & \citeyear{Cao2018Randomly} & \cref{sec:rtprelu} & \cite{Cao2018Randomly} & \cmark &  $a$  &  $\alpha$ &  controllable slope only for certain inputs; stochastic thresholding \\ 
        \gls{PREU} & \citeyear{Ying2019} & \cref{sec:preu} & \cite{Ying2019} & \cmark & $a$, $b$  &  $\alpha$, $\beta$ & horizontal scaling only for negative inputs  \\
        \gls{AReLU} & \citeyear{Chen2020AReLU} & \cref{sec:arelu} & \cite{Chen2020AReLU} & \cmark & $a_l$, $b_l$   & $\alpha$  &  separate scaling for negative and positive inputs; parameter transformation   \\
        \gls{tanhLU} & \citeyear{Shen2022} & \cref{sec:tanhlu} & \cite{Shen2022} & \cmark & $a_i$, $b_i$, $c_i$   & $\alpha$, $\beta$  &  separate scaling for each component \\
        \gls{DPReLU} & \citeyear{Balaji2020}, \citeyear{Varshney2021} & \cref{sec:dprelu} & \cite{Balaji2020,Varshney2021} & \cmark & $a_i$, $b_i$   &  $\alpha$ & proposed independently in \cite{Balaji2020} and \cite{Varshney2021}  \\
        \gls{dual_line} & \citeyear{Balaji2020} & \cref{sec:dual_line} & \cite{Balaji2020} & \cmark & $a_i$, $b_i$, $m_i$   &  $\alpha$, $\delta$  & separate scaling for negative and positive inputs; common vertical translation \\
        \gls{PiLU} & \citeyear{Inturrisi2021} & \cref{sec:pilu} & \cite{Inturrisi2021} & \cmark & $a_i$, $b_i$, $c_i$   &  $\alpha$, $\gamma$  & separate scaling for negative and positive inputs; common horizontal translation \\
        \gls{DPAF} & \citeyear{Balaji2020} & \cref{sec:dpaf} & \cite{Balaji2020} & \cmark &  $a_i$, $m_i$ &  $\alpha$, $\delta$  &  slope scaling only for positive inputs; common vertical translation  \\
        \gls{FPAF} & \citeyear{Varshney2021} & \cref{sec:fpaf} & \cite{Varshney2021} & \cmark &  v &  $\alpha$  & separate scaling for negative and positive inputs \\
        \gls{EPReLU} & \citeyear{Jiang2018} & \cref{sec:eprelu} & \cite{Jiang2018} & \cmark &  $a_i$, $k_i$   &  $\alpha$ &  separate scaling for negative and positive inputs, stochastic scaling for positive inputs \\
        \gls{PTELU} & \citeyear{Gupta2017} & \cref{sec:ptelu} & \cite{Gupta2017} & \cmark & $a_i$, $b_i$   &  $\alpha$, $\beta$  & scaling for negative inputs only  \\
        \gls{TReLU} & \citeyear{Zhang2019Research} & \cref{sec:trelu} & \cite{Zhang2019Research} & \cmark &  $b_i$  &  $\beta$  & scaling for negative inputs only    \\
        \gls{BLU} & \citeyear{Godfrey2019} & \cref{sec:blu} & \cite{Godfrey2019} & \cmark &  $a_i$  &  $\alpha$ & scaling only the nonlinear component  \\
        \gls{PELU} & \citeyear{Trottier2016} & \cref{sec:pelu} & \cite{Trottier2016} & \cmark & $a_i$, $b_i$   & $\alpha$, $\beta$  &  separate vertical scaling for positive and negative inputs, horizontal scaling only for negative inputs \\
        \gls{FELU} & \citeyear{Qiumei2019} & \cref{sec:felu} & \cite{Qiumei2019} & \cmark & $a_i$   &  $\alpha$  &  scaling only for negative inputs  \\
        \gls{MPELU} & \citeyear{Li2018Improving} & \cref{sec:mpelu} & \cite{Li2018Improving} & \cmark &  $a_i$, $b_i$  &  $\alpha$, $\beta$  &  scaling only for negative inputs  \\
        \gls{CELU} & \citeyear{Barron2017} & \cref{sec:celu} & \cite{Barron2017} & \cmark &  $a_i$  & $\alpha$, $\beta$  &  scaling only for negative inputs, continuously diff.\\
        \gls{PSELU} & \citeyear{Pratama2020} & \cref{sec:pselu} & \cite{Pratama2020} & \cmark &  $a_i$, $b_i$  & $\alpha$  & separate vertical scaling for positive and negative inputs  \\
        \gls{LPSELU} & \citeyear{Pratama2020} & \cref{sec:lpselu} & \cite{Pratama2020} & \cmark & $a_i$, $b_i$, $c_i$   & $\alpha$  & individual components have separate parameters for controlling the slope \\
        \gls{LPSELURP} & \citeyear{Pratama2020} & \cref{sec:lpselurp} & \cite{Pratama2020} & \cmark & $a_i$, $b_i$, $c_i$, $m_i$   & $\alpha$, $\delta$  & individual components have separate parameters for controlling the slope  \\
        \gls{PDELU} & \citeyear{Cheng2020} & \cref{sec:pdelu} & \cite{Cheng2020} & \cmark &  $a_i$, $b$  & $\alpha$   & scaling only for negative inputs, fixed $b$ for shape control \\
        \gls{T-swish} & \citeyear{Javid2022} & \cref{sec:tswish} & \cite{Javid2022} & \cmark &  $a_i$, $b_i$, $c_i$  & $\alpha$, $\beta$  & scaling only for negative inputs, $c_i$ for threshold determination\\
        \gls{EELU} & \citeyear{Kim2020} & \cref{sec:eelu} & \cite{Kim2020} & \cmark & $a^c$, $b^c$, $k_i^c$   &  $\alpha$, $\beta$  & vertical and horizontal scaling for negative inputs; vertical stochastic scaling for positive inputs \\
        \gls{scaled_softsign} & \citeyear{Pishchik2023} & \cref{sec:scaled_softsign} & \cite{Pishchik2023} & \cmark &  $a_i$, $b_i$  & $\alpha$  &  additional adaptive parameter  \\
        \gls{NAF} & \citeyear{ShuxiangXu2000} & \cref{sec:naf} & \cite{ShuxiangXu2000} & \cmark & $a$, $b$, $c$, $d$   &  $\alpha$, $\beta$  & each component has its own scaling  \\
        \gls{SLS-SS} & \citeyear{Tezel2007} & \cref{sec:scaled_logistic_sigmoid} & \cite{Tezel2007} & \cmark & $a_i$, $b_i$, $c_i$, $d_i$    & $\alpha$, $\beta$  & each component has its own scaling  \\
        \gls{APLU} & \citeyear{Hou2017} & \cref{sec:aplu} & \cite{Hou2017} & \cmark & $a_i^s$, $b_i^s$, $s=1,\ldots,S$   & $\alpha$, $\gamma$  & sum of $S$ \glspl{TAAF}  \\
        \gls{SReLU} & \citeyear{Jin2016} & \cref{sec:srelu} & \cite{Jin2016} & \cmark & $t_i^r$, $a_i^r$, $t_i^l$, $a_i^l$   & $\alpha$  & slope controllable only for the outermost segments  \\
        \gls{LinQ} & \citeyear{Bilski2016} & \cref{sec:linq} & \cite{Bilski2016} & \cmark & $a$  &  $\alpha$  & slope controllable only for the outermost segments outside the interval $[-2,2]$  \\
        \gls{All-ReLU} & \citeyear{Curci2021} & \cref{sec:allrelu} & \cite{Curci2021} & \xmark & $a$   & $\alpha$  & slope controllable only for negative inputs, alternating between layers  \\
        \gls{PLU} & \citeyear{Nicolae2018} & \cref{sec:plu} & \cite{Nicolae2018} & \cmark &  $a_i$, $b$  &  $\alpha$ & $b$ fixed;   slope controllable only for the outermost segments  \\
        \gls{AdaLU} & \citeyear{Mo2022} & \cref{sec:adalu} & \cite{Mo2022} & \cmark &  $a_i$, $b_i$, $c_i$, $d_i$,  $e_i$  &  $\alpha$, $\gamma$, $\delta$& $b$ controllable slopes and offsets for individual components\\
        \gls{MTLU} & \citeyear{Gu2019} & \cref{sec:mtlu} & \cite{Gu2019} & \cmark &  $a_{i,0}, \ldots, a_{i, K}$, $b_{i,0}, \ldots, b_{i, K}$, $c_{i,0}, \ldots, c_{i, K-1}$  & $\alpha$, $\delta$  &  separate parameters for indiv. segments  \\
        \gls{maxout} & \citeyear{Goodfellow2013} & \cref{sec:maxout} & \cite{Goodfellow2013} & \cmark &  $w_i^k$, $b_i^k$, $k=1,\ldots,K$  & $\alpha$, $\delta$  &  maximum of individually transformed functions  \\
        \gls{DY--ReLU} & \citeyear{Chen2020Dynamic} & \cref{sec:dyrelu} & \cite{Chen2020Dynamic} & \cmark &  $a_{i,k}$, $b_{i,k}$, $k=1,\ldots,K$  &   $\alpha$, $\delta$  &  maximum of individually transformed functions; hyperfunction for parameter optimization  \\
        \gls{ABU} & \citeyear{Stfeld2020} & \cref{sec:abu} & \cite{Stfeld2020} & \cmark & $a_{i,j}$   &  $\alpha$  & sum of individually transformed functions   \\
        \gls{ABU} with bias & \citeyear{Wang2018LookUp} & \cref{sec:abu} & \cite{Wang2018LookUp} & \cmark & $a_{i,j}$, $b_{i,j}$   &  $\alpha$, $\beta$  & sum of individually transformed functions   \\
        \gls{ABU} (constrained) & \citeyear{Manessi2018} & \cref{sec:abu} & \cite{Manessi2018} & \cmark & $a_{i,j}$  &  $\alpha$ & sum of individually transformed functions; their scaling parameter sum up to 1   \\
        \gls{TCA} & \citeyear{Baggenstoss2022} & \cref{sec:tca} & \cite{Baggenstoss2022} & \cmark & $a_{i,j}$, $b_{i,j}$   &  $\beta$, $\gamma$  & sum of individually transformed functions   \\
        \gls{TCAv2} & \citeyear{Baggenstoss2023} & \cref{sec:tca} & \cite{Baggenstoss2023} & \cmark & $a_{i,j}$, $b_{i,j}$, $c_{i,j}$   &  $\alpha$, $\beta$, $\gamma$  & sum of individually transformed functions   \\
        activation ensemble & \citeyear{Klabjan2019} & \cref{sec:abu} & \cite{Klabjan2019} & \cmark & $a_{i,j}$  &  $\alpha$ & sum of individually transformed functions; their scaling parameter sum up to 1; min--max scaling   \\
        \gls{SinLU} & \citeyear{Gupta2017} & \cref{sec:sinlu} & \cite{Paul2022} & \cmark & $a_i$, $b_i$   &  $\alpha$, $\beta$  & scaling only of a single term  \\
        \gls{GABU}  & \citeyear{Dushkoff2016} & \cref{sec:gabu} & \cite{Dushkoff2016} & \cmark &  $a_{i,j}$  &  $\alpha$   & sum of individually transformed functions; gated  \\
        \gls{SLAF}  & \citeyear{Goyal2019} & \cref{sec:slaf} & \cite{Goyal2019} & \cmark &  $a_{i,j}$  &  $\alpha$  &  sum of individually transformed functions \\
        \gls{ChPAF}  & \citeyear{Deepthi2023} & \cref{sec:chpaf} & \cite{Deepthi2023} & \cmark & $a_j$, $j=0,\ldots,k$   &    $\alpha$  &  sum of individually transformed functions \\
        \gls{LPAF} & \citeyear{Venkatappareddy2021} & \cref{sec:lpaf} & \cite{Venkatappareddy2021} & \cmark & $a_j$, $j=0,\ldots,k$   &   $\alpha$  &  sum of individually transformed functions \\
        \gls{KAF} & \citeyear{Scardapane2019} & \cref{sec:kaf} & \cite{Scardapane2019} & \cmark &  $a_{i,j}$, $d_j$, $j=1,\ldots,D$ & $\alpha$  & $d_j$ fixed; sum of individually transformed functions  \\
        \gls{PAU}  & \citeyear{Molina2020} & \cref{sec:pau} & \cite{Molina2020} & \cmark &  $a_{j}$, $j=0,\ldots,m$, $b_{k}$, $k=1,\ldots,n$  & $\alpha$   & division of two sums of sum of individually transformed functions \\
        \gls{MoGU}  & \citeyear{Wang2018LookUp} & \cref{sec:mogu} & \cite{Wang2018LookUp} & \cmark &  $a_{i,j}$, $\sigma_{i,j}$, $\mu_{i,j}$, $j=1,\ldots,n$ & $\alpha$, $\beta$, $\gamma$  &  sum of individually transformed functions \\
        \gls{MSAF}  & \citeyear{Cai2015} & \cref{sec:msaf} & \cite{Cai2015} & \xmark &  $a$, $b_k$, $k=1,\ldots,N$ & $\gamma$, $\delta$  &  sum of individually translated functions  \\
        \gls{FSA}  & \citeyear{Liao2020} & \cref{sec:fsa} & \cite{Liao2020} & \cmark & $a_i$, $b_{i,j}$, $c_{i,j}$, $d_i$, $j=0,\ldots,r$   & $\alpha$, $\beta$, $\delta$   & sum of individually transformed functions  \\
        \gls{VAF}  & \citeyear{Apicella2019} & \cref{sec:vaf} & \cite{Apicella2019} & \cmark & $a_{l,0}$, $a_{l,j}$, $b_{l,j}$, $c_{l,j}$, $j=1, \ldots, k$   &  $\alpha$, $\beta$, $\gamma$, $\delta$  &  sum of \glspl{TAAF}  \\
        \hline
        \caption[Activation functions related to TAAFs]{\textbf{Activation functions related to \glspl{TAAF}} \\ \Glsxtrlongpl{AF} that employs the same or similar concepts as \glspl{TAAF} as listed in \cite{Kunc2024Decades}. The column \textit{TAAF equiv.} lists the \gls{TAAF}'s parameters whose function the activation also employs in any manner.}
        \label{tab:act_related_activations}
    \end{longtable}
 \end{landscape}